\documentclass[11pt]{article}

\usepackage{acl}

\usepackage{times}
\usepackage{latexsym}
\usepackage[T1]{fontenc}
\usepackage[utf8]{inputenc}
\usepackage{microtype}
\usepackage{inconsolata}
\usepackage{booktabs}
\usepackage{tabularx}
\usepackage{enumitem}
\usepackage{graphicx}
\usepackage{amsmath}
\usepackage{subcaption} 
\usepackage{ragged2e}       
\usepackage{array}
\usepackage{multirow}       
\usepackage{threeparttable} 
\usepackage{enumitem}       
\usepackage[table,xcdraw]{xcolor} 
\usepackage{placeins} 

\usepackage[most]{tcolorbox}
\tcbuselibrary{skins, breakable}
\usepackage{listings}

\providecommand{\tightlist}{%
  \setlength{\itemsep}{0pt}\setlength{\parskip}{0pt}}

\newcommand\blfootnote[1]{%
  \begingroup
  \renewcommand\thefootnote{}\footnote{#1}%
  \addtocounter{footnote}{-1}%
  \endgroup
}

\newtcolorbox{greybox}{
    colback=gray!10,
    colframe=black,
    width=\linewidth,
    boxrule=0.5pt,
    enhanced,
    breakable,
    before skip=10pt,
    after skip=10pt,
    top=5pt,
    bottom=5pt,
    left=2pt,
    right=2pt,
    fontupper=\raggedright
}

\lstdefinestyle{sqlstyle}{
    language=SQL,
    backgroundcolor=\color{black!5},   
    commentstyle=\color{green!50!black},
    keywordstyle=\color{blue}\bfseries,
    stringstyle=\color{red!80!black},
    numberstyle=\tiny\color{gray},
    basicstyle=\ttfamily\small,        
    breakatwhitespace=false,         
    breaklines=true,                 
    captionpos=b,                    
    keepspaces=true,                 
    numbers=none,                    
    numbersep=5pt,                   
    showspaces=false,                
    showstringspaces=false,
    showtabs=false,
    frame=single,                    
    rulecolor=\color{black!40},
    tabsize=2
}

\title{LOGIGEN: Logic-Driven Generation of Verifiable Agentic Tasks}

\author{
\textbf{Yucheng Zeng\textsuperscript{1,*}}\quad
\textbf{Weipeng Lu\textsuperscript{1,*}}\quad
\textbf{Linyun Liu\textsuperscript{1,*}}\quad
\textbf{Shupeng Li\textsuperscript{1,*,$\dagger$,$\ddagger$}}\quad 
\textbf{Zitian Qu\textsuperscript{2}}\quad 
\textbf{Chenghao Zhu\textsuperscript{1}}\quad \\
\textbf{Shaofei Li\textsuperscript{1}}\quad
\textbf{Zhengdong Tan\textsuperscript{1}}\quad
\textbf{Mengyue Liu\textsuperscript{1}}\quad
\textbf{Haotian Zhao\textsuperscript{1}}\quad
\textbf{Zhe Zhou\textsuperscript{2}}\quad
\textbf{Jianmin Wu\textsuperscript{1,$\dagger$}}\\
\\
\textsuperscript{1} Baidu Inc. \quad
\textsuperscript{2} Tsinghua University \\
\\
\texttt{\{zengyucheng, luweipeng, liulinyun, lishupeng, zhuchenghao, } \\
\texttt{lishaofei01, tanzhendong, liumengyue, zhaohaotian02, wujianmin\}@baidu.com} \\
\texttt{\{qzt22, z-zhou24\}@mails.tsinghua.edu.cn} \\
}

\begin{document}
\maketitle
\blfootnote{* Equal contributors. $\dagger$ Corresponding authors. $\ddagger$ Project leader.}
\begin{abstract}

The evolution of Large Language Models (LLMs) from static instruction-followers to autonomous agents necessitates operating within complex, stateful environments to achieve precise state-transition objectives. However, this paradigm is bottlenecked by data scarcity, as existing tool-centric reverse-synthesis pipelines fail to capture the rigorous logic of real-world applications. We introduce \textbf{LOGIGEN}, a logic-driven framework that synthesizes verifiable training data based on three core pillars: \textbf{Hard-Compiled Policy Grounding}, \textbf{Logic-Driven Forward Synthesis}, and \textbf{Deterministic State Verification}. Specifically, a Triple-Agent Orchestration is employed: the \textbf{Architect} compiles natural-language policy into database constraints to enforce hard rules; the \textbf{Set Designer} initializes boundary-adjacent states to trigger critical policy conflicts; and the \textbf{Explorer} searches this environment to discover causal solution paths. This framework yields a dataset of 20,000 complex tasks across 8 domains, where validity is strictly guaranteed by checking exact state equivalence. Furthermore, we propose a verification-based training protocol where Supervised Fine-Tuning (SFT) on verifiable trajectories establishes compliance with hard-compiled policy, while Reinforcement Learning (RL) guided by dense state-rewards refines long-horizon goal achievement. On $\tau^2$-Bench, LOGIGEN-32B(RL) achieves a \textbf{79.5\% success rate}, substantially outperforming the base model (40.7\%). These results demonstrate that logic-driven synthesis combined with verification-based training effectively constructs the causally valid trajectories needed for next-generation agents.

\end{abstract}

\begin{figure*}[h]
    \centering
    \begin{subfigure}[b]{0.74\textwidth}
        \centering
        \includegraphics[width=\textwidth]{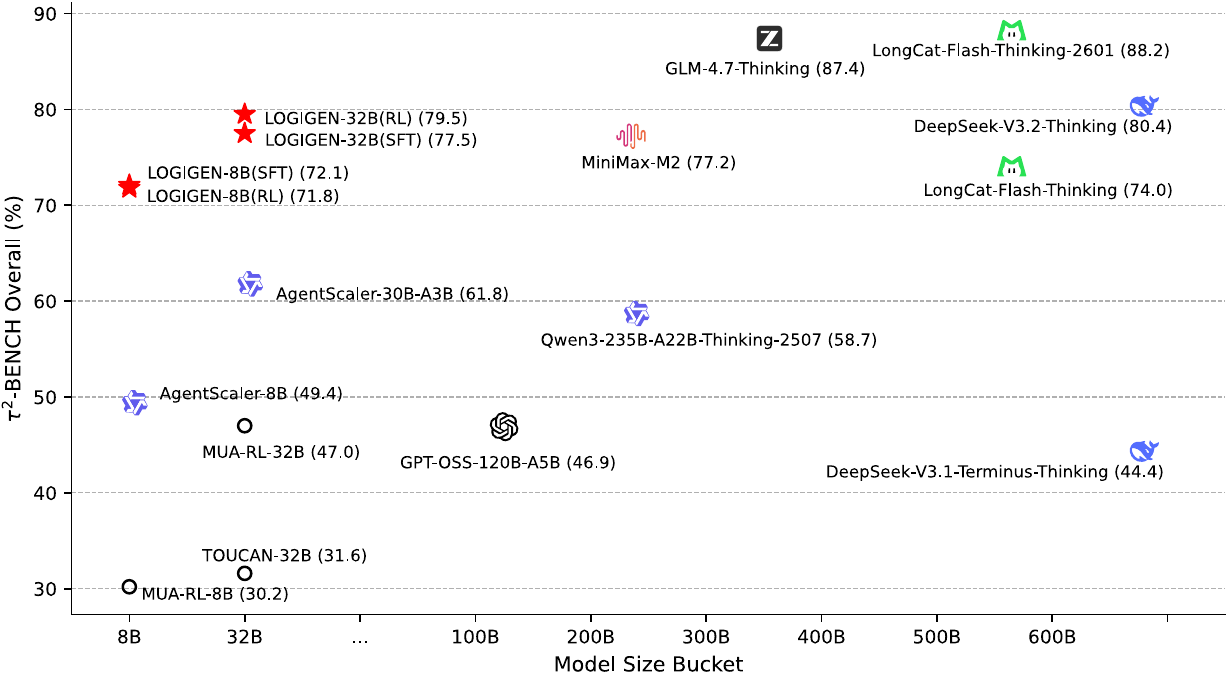}
        \caption{Performance comparisons.}
        \label{fig:perf_comparisons}
    \end{subfigure}
    \hfill
    \begin{subfigure}[b]{0.25\textwidth}
        \centering
        \includegraphics[width=\textwidth]{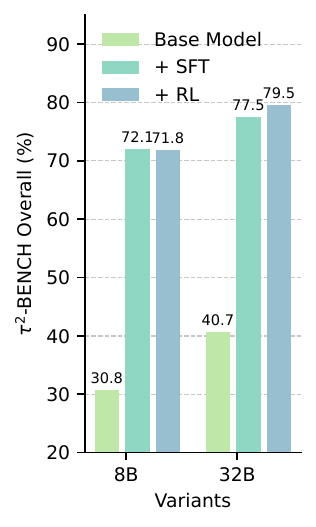}
        \caption{Performance breakdown.}
        \label{fig:incre_gains}
    \end{subfigure}
    \caption{
    \textbf{(a) Performance comparisons.} LOGIGEN-32B (RL) achieves a 79.5\% success rate on $\tau^2$-Bench, significantly outperforming open-weight baselines and remaining competitive with proprietary models. It also surpasses general-purpose baselines with substantially larger parameters.
    \textbf{(b) Performance breakdown.} The gains stem from our verification-based training protocol: SFT on verifiable trajectories establishes compliance with hard-compiled policy, while RL guided by deterministic state-rewards refines long-horizon goal achievement.
    }
    \label{fig:into_overall}
\end{figure*}

\section{Introduction}
\label{sec:introduction}

The evolution of Large Language Models (LLMs) from static instruction-followers to autonomous agents marks a fundamental shift in the AI landscape. Existing evaluation frameworks, such as BFCL~\citep{bfcl} and ACE-Bench~\citep{acebench}, focus primarily on syntactic tool-mapping, which evaluates whether a model can correctly generate tool names and parameters based on an isolated query without actual execution. For instance, these benchmarks typically test if a model can map a query like ``What is the weather in New York?'' to a predefined function call \texttt{get\_weather(city="New York")}, verifying only the syntax rather than the execution outcome. This instruction-response paradigm treats function calls as isolated events, effectively decoupling tool usage from environmental execution and ignoring the stateful dependencies essential for agentic workflows. In contrast, real-world agents must operate within a stateful environment (e.g., a database). Here, the system state evolves dynamically: for example, a payment action updates the account balance, and this new state immediately determines whether subsequent transactions are permitted or rejected based on wiki policy (domain-specific operational rules). To address these complexities, the $\tau$-Bench family~\citep{tau1, tau2} introduces a more rigorous evaluation framework. It defines task success as a deterministic state transition objective: the model must transform an initial environment state $s_{\text{origin}}$ into a specific target state $s_{\text{target}}$, ensuring that all intermediate steps comply with the underlying logical constraints.

At its core, this transition reflects a fundamental paradigm shift from the static “Instruction-Response” to the dynamic “Action-Observation”\citep{era, react}. Under this new paradigm, as posited in Era of Experience\citep{era}, an agent's growth depends on continuous action-observation loops where the environment acts as a strict validator. This mechanism forces the agent to learn from complex feedback chains, such as attempting actions, encountering constraints, and refining plans to achieve success. However, relying on such environmental interaction creates a dual bottleneck: unlike the straightforward curation of static instruction pairs, data collection now necessitates active exploration to discover causally valid paths, while reliable supervision demands objective rewards derived directly from environmental states. Consequently, traditional data construction methods fall short, as they inherently lack the interactive exploration process and deterministic feedback signals essential for this paradigm.

While $\tau$-Bench provides a rigorous evaluation benchmark, the community currently lacks the scalable training data necessary to cultivate agents capable of mastering its complex state-transition tasks. Existing synthesis pipelines~\citep{xlam, toolace, toucan, agentscaler} attempt to bypass the need for a stateful execution environment through \textbf{tool-centric reverse-synthesis}. Typically, these methods start from an observed tool sequence (e.g., \texttt{check\_balance} $\to$ \texttt{transfer}) and retroactively generate a user query (e.g., \textit{``Pay my bill''}). This approach fundamentally fails to capture the essence of the Action-Observation paradigm for three reasons: (i) \textbf{Decoupling from Execution Feedback}: By generating trajectories in a static vacuum, models mimic the surface form of tool use without experiencing the causal feedback loops involving error messages and state updates that are essential for real-world decision-making; (ii) \textbf{Bias toward Happy Paths}: Since reverse-synthesis relies on pre-defined successful chains, it inherently under-samples boundary interactions, ignoring the hard constraints (e.g., insufficient funds) that require agents to negotiate or recover; and (iii) \textbf{Lack of State-Based Verification}: It provides only textual supervision, lacking the deterministic environmental states required to rigorously verify whether the task was actually completed.

In this paper, we argue that a rigorous agentic training set must be constructed upon three core pillars that directly address these deficiencies:
\begin{enumerate}
\item \textbf{Hard-Compiled Policy Grounding}: To address the \textbf{decoupling from execution feedback}, wiki policy must be hard-compiled into the execution environment. This ensures that the system provides deterministic feedback when constraints are violated, forcing the agent to learn from boundary failures rather than merely replicating smooth success stories.
\item \textbf{Logic-Driven Forward Synthesis}: To counteract the \textbf{bias toward happy paths}, tasks must be synthesized via logical exploration. By deductively deriving valid paths from an initial state, this approach preserves the causal integrity of the action-observation loop, ensuring every step is a logical consequence of the environmental state rather than a retroactive reconstruction.
\item \textbf{Deterministic State Verification}: To resolve the \textbf{lack of state-based verification}, success must be measured by computing the ``State-Diff'', which checks whether the final environment state exactly matches the ground truth target. This verification provides an objective reward signal essential for reliable reinforcement learning.
\end{enumerate}

To implement these principles, we introduce \textbf{LOGIGEN}, a logic-driven framework for autonomous data synthesis. By compiling wiki policy into a stateful execution environment, LOGIGEN moves beyond \textbf{reverse-synthesis} to \textbf{deductive generation}, where trajectories are derived directly from the hard constraints of business logic. The framework employs a \textbf{Triple-Agent Orchestration} to automate this deductive process: the \textit{Architect} compiles natural-language wiki policy into execution environment constraints (comprising database tables and triggers); the \textit{Set Designer} initializes boundary-adjacent states where policy conflicts are most likely to occur; and the \textit{Explorer} searches this strictly constrained environment to discover valid causal paths through trial and error. This approach ensures that every synthesized trajectory is not only grounded in this stateful environment but also logically rigorous with respect to complex wiki policy.

Empirical results validate this architectural shift. As illustrated in Figure~\ref{fig:into_overall}, our LOGIGEN-32B(RL) model achieves a \textbf{79.5\% success rate} on $\tau^2$-Bench, significantly outperforming both same-sized agent frameworks and general-purpose models with substantially larger parameters. This performance is driven by the synergy between logic-driven data synthesis and a verification-based training protocol: SFT on verifiable trajectories establishes compliance with hard-compiled policy, while RL guided by deterministic state-rewards refines long-horizon goal achievement.

Our contributions are summarized as follows:
\begin{itemize}
\item \textbf{Principled Framework for Agentic Data}: We identify three essential principles for constructing rigorous agent training data: \textit{Hard-Compiled Policy Grounding} for deterministic feedback, \textit{Logic-Driven Forward Synthesis} for causal validity, and \textit{Deterministic State Verification} for objective supervision. This framework provides a theoretical foundation to address the deficiencies of existing tool-centric reverse-synthesis pipelines.
\item \textbf{Scalable Instantiation via Triple-Agent Orchestration}: We realize these principles in LOGIGEN, an automated framework featuring a Triple-Agent Orchestration (Architect, Set Designer, Explorer). This system successfully synthesizes a dataset of 20,000 logic-dense tasks across diverse domains, effectively resolving the bottleneck of scalable training data for complex state-transition objectives.
\item \textbf{State-of-the-Art Empirical Performance}: We validate our approach on $\tau^2-Bench$, where LOGIGEN-32B(RL) achieves a 79.5\% success rate, significantly surpassing open-weight baselines and remaining competitive with proprietary models. This demonstrates the efficacy of combining logic-driven synthesis with verification-based training protocols (SFT + RL).
\end{itemize}

\newcolumntype{Y}{>{\RaggedRight\arraybackslash\small}X}
\newcolumntype{A}{>{\hsize=0.68\hsize}Y} 
\newcolumntype{B}{>{\hsize=0.87\hsize}Y} 
\newcolumntype{C}{>{\hsize=1.35\hsize}Y} 
\begin{table*}[t]
\centering
\setlength{\tabcolsep}{3pt} 
\renewcommand{\arraystretch}{1.25}

\begin{threeparttable}
    \caption{\textbf{Comparison from a dataset construction perspective.} 
    \emph{Hard-compiled} refers to policies enforced by database constraints.
    \emph{Deterministic state check} ensures correctness via reproducible state comparisons.}
    \label{tab:construction-compare}

    \begin{tabularx}{\textwidth}{@{} A B B B C C @{}}
    \toprule
    \textbf{Dataset} &
    \textbf{Policy \newline Encoding} & 
    \textbf{Env. \newline Grounding} &
    \textbf{Init. \newline Sampling} &
    \textbf{Task \newline Synthesis} &
    \textbf{Verification} \\
    \midrule

    API-Bank  \newline ToolACE \newline TOUCAN &
    Tool/API/MCP specs \newline (no explicit policy) &
    Mocked or external tool execution \newline (no released local stateful env) & 
    N/A \newline (Stateless) &
    Spec-conditioned \newline LLM task synthesis \newline (tool-first / reverse) \newline \textbf{(Training Data)} & 
    Rule / metric checks \newline (Parsers / regex) \newline (no state-diff) \\
    \addlinespace

    EnvScaler \newline AgentScaler &
    Programmatic logic \tnote{a} &
    Executable simulator &
    Programmatic \newline (state synthesis) &
    Tool-graph / scenario-driven \newline (tool-first / reverse) \newline \textbf{(Training Data)} &
    State-based checks \newline (validators / alignment)\tnote{b} \\
    \addlinespace

    $\tau$-Bench \newline $\tau^2$-Bench &
    Soft \newline (NL guidelines + env rules) &
    Executable env \newline w/ stateful backend &
    Curated task-provided init state &
    Human-authored tasks \newline \textbf{(Benchmark)} &
    Deterministic state check \\
    \addlinespace

    \rowcolor{gray!15}
    \textbf{LOGIGEN (Ours)} &
    \textbf{Hard-Compiled Policy} &
    \textbf{Executable DB-backed env} &
    \textbf{Boundary-adjacent initial states} &
    \textbf{Automated logic-driven} \newline \textbf{(Training Data)} &
    \textbf{Deterministic state check} \\
    \bottomrule
    \end{tabularx}

    \begin{tablenotes}[flushleft]
    \scriptsize
    \item[a] indicates construction via code validators rather than NL-only policies.
    \item[b] Denotes state-based validation/alignment, potentially supplemented by LLM judges.
    \item \textit{Note:} We include representative frameworks; the list is not exhaustive.
    \end{tablenotes}
\end{threeparttable}
\end{table*}

\section{Related Work}

The development of capable tool-use agents has advanced along three interconnected directions: (i) \emph{benchmarks} that define success criteria for real-world tool-use tasks, (ii) \emph{data synthesis} methods that scale training data beyond limited human annotation, and (iii) \emph{training and evaluation loops} that provide reliable feedback for achieving long-horizon goals. To understand the bottlenecks limiting progress in these areas, Table~\ref{tab:construction-compare} analyzes representative work through the lens of dataset construction. It highlights four recurring design choices: \textbf{policy encoding} (schema-only vs.\ enforceable constraints), \textbf{environment grounding} (mocked vs.\ executable), \textbf{task synthesis} (reverse-synthesis vs.\ logic-driven), and \textbf{verification} (rule checks vs.\ deterministic state checks). We organize this section around these dimensions to clarify what current pipelines provide---and, crucially, what remains missing for agents to master policy-governed state transitions.

\subsection{From Atomic Function Calling to State-Based Evaluation}
Early tool-use research often framed tool interaction as \textbf{atomic function calling}: given an instruction, the model generates a syntactically valid function name and parameters~\citep{apibank, gorilla, toolbench, react}. Benchmarks such as ACE-Bench~\citep{acebench} and BFCL~\citep{bfcl} further standardized evaluation formats by prioritizing \textbf{syntactic adherence}—verifying that function names and parameters strictly match the predefined ground truth, rather than the semantic correctness of the execution outcome. While foundational, these settings are typically \textbf{stateless} (or weakly stateful): success is determined by the validity of a single, isolated function-calling, rather than by the successful completion of a \textbf{multi-step workflow} where each action dynamically alters the system state and constraints for subsequent steps. This paradigm implicitly encourages agents to behave as mere \textbf{translators} that map natural language to tool syntax, rather than \textbf{decision makers} capable of navigating trade-offs, such as resolving a user's urgent request while strictly enforcing a non-refundable policy. Consequently, while effective for assessing a model's capacity to generate syntactically correct tool invocations, these benchmarks fail to evaluate the causal reasoning and long-horizon planning required in policy-rich workflows.

To address these limitations, the $\tau$-Bench family~\citep{tau1, tau2} shifts the evaluation paradigm by defining success as a \textbf{deterministic state transition objective}. Specifically, it requires agents to transform an initial environment state $s_{\text{origin}}$ into a ground truth target state $s_{\text{target}}$, ensuring that all intermediate steps strictly adhere to logical constraints. This framework establishes a rigorous standard for testing agentic competence in policy-rich environments. However, a critical disconnect remains: while the benchmark provides a comprehensive \textit{test}, the community lacks the scalable \textit{training data} necessary for agents to master it. Existing pipelines struggle to generate diverse, policy-consistent data that match the complexity of these state-based evaluations, creating a bottleneck that motivates research into environment-grounded synthesis and verifiable training loops.

\subsection{Tool-Centric Reverse Synthesis}

A dominant paradigm for scaling tool-use data is \textbf{tool-centric reverse-synthesis}~\citep{xlam, apigen, toolace, toolcoder, toucan, agentscaler, envscaler}. These pipelines typically start from pre-defined tool schemas or observed execution traces (e.g., sequences of tool calls) and prompt an LLM to retroactively generate user queries and dialogues that \textbf{correspond to} these tool sequences. While attractive for its scalability, this approach treats the \emph{execution trace} itself as the implicit learning objective: models learn to imitate surface-level patterns rather than to reason about the causal structure of policy-governed state transitions. As a result, the synthesized data often exhibit low ``logical density'': they may consist of smooth, uninterrupted dialogues, yet contain few genuine decision points where user intent collides with system constraints (e.g., failed transactions due to quota limits or approval requirements), which are central to real-world applications.

Moreover, because reverse-synthesis is commonly conditioned on \emph{successful} execution traces (``happy paths''), it systematically under-samples boundary cases where policies reject actions and agents must recover via clarification, negotiation, or alternative plans. This limitation is particularly problematic, as the most informative supervision signals for learning robust agency often arise from state-dependent failures and subsequent corrective actions, rather than from smooth, uninterrupted execution.

\subsection{Execution Environments and Policy Embedding}
To mitigate the brittleness of reverse-synthesis methods that rely solely on tool schemas or execution traces, recent works have adopted \textbf{execution environments} that validate actions through executable simulators~\citep{envscaler, agentscaler, LongCat-Flash-Thinking-2601}. Unlike static LLM-mocked responses, executable simulators rely on programmatic logic (i.e., executing actual code rather than generating text), thereby exposing agents to genuine return values, realistic error codes, and the actual side effects of function calls. 

While this confirms the importance of grounding agents in executable environments~\citep{era}, a critical limitation remains: many existing pipelines treat the environment merely as a sandbox for executing \textbf{isolated function calls}. In these systems, the \textbf{functions, data, and policies exist in silos}, lacking a systemic binding. They rely on external simulators (e.g., Python scripts) where policy enforcement depends on \textbf{imperative programmatic checks} applied at the function level. This decoupling means that business rules are not inextricably linked to the data model or the function logic. Consequently, policy enforcement becomes \textbf{brittle}: if a specific check is omitted or inconsistent, the environment may permit erroneous state changes, failing to provide the rigorous, invariant guarantees required for training robust agents.

In contrast, our approach compiles policy directly into database triggers, establishing \textbf{atomic, declarative constraints}. This binds the rules to the data model, ensuring that the database engine enforces them at the \textbf{transactional level} for every write operation. This distinction is fundamental: it moves policy enforcement from a code-level heuristic to a system-level invariant. Invalid actions are physically rejected by the database engine, ensuring that synthesized data are forged through the hard collision between agent intent and \textbf{inescapable physical boundaries}, rather than merely simulated safeguards.

\subsection{Deterministic State Verification and Reward}

Enhancing agentic competence relies on effective training, which fundamentally demands accurate \textbf{reward signals}, particularly for reinforcement learning (RL) over long-horizon tasks. Optimizing via \textbf{deterministic state-rewards} has been explored in prior RL work~\citep{mua-rl}, where optimizing for the final target state outperforms process-oriented supervision (e.g., rule matching or trajectory imitation). However, the effectiveness of such training is tightly coupled to the verifiability of the reward signal.

A common practice is to rely on rule matching (e.g., checking whether a target function call appears) or on \textbf{LLM-as-a-Judge} style models to generate these rewards~\citep{g-eval, judginllm-as-a-judge, surveyllmasajudge, judgingjudgessystematicstudy, trustllmjudgmentsreliability, trustjudge}. While convenient, judge-based reward signals can be noisy and susceptible to reward hacking~\citep{toolrl, toolrm, proofofuse}: agents may learn to generate convincing conversations without reliably inducing the correct underlying state transition. $\tau$-Bench family ~\citep{tau1, tau2} highlights an alternative: \textbf{deterministic state checks} that directly compare the final environment state against a ground-truth target. This approach aligns with real-world operational objectives, where success is measured by achieving the predetermined state, rather than generating convincing conversations. This serves as a natural foundation for scalable RL, crucially allowing us to determine whether the rollout trajectory truly teaches agents to satisfy policy-governed constraints rather than merely optimizing for superficial textual plausibility.

In summary, while prior work has made substantial progress, a foundational gap persists. Existing pipelines lack a unified framework that satisfies three essential properties for rigorous agent training (Table~\ref{tab:construction-compare}): 
(i) Hard-Compiled Policy Grounding via non-negotiable, compiled rules,
(ii) Logic-Driven synthesis ensuring data are discovered from environmental logic rather than reconstructed from observed execution traces, 
and (iii) Deterministic Verification to enable objective rewards for optimization. 
Our work is the first to jointly address these three requirements, treating policy-governed state transitions as the core object of both data generation and evaluation.

\begin{figure}[t]
  \centering
  \includegraphics[width=\linewidth]{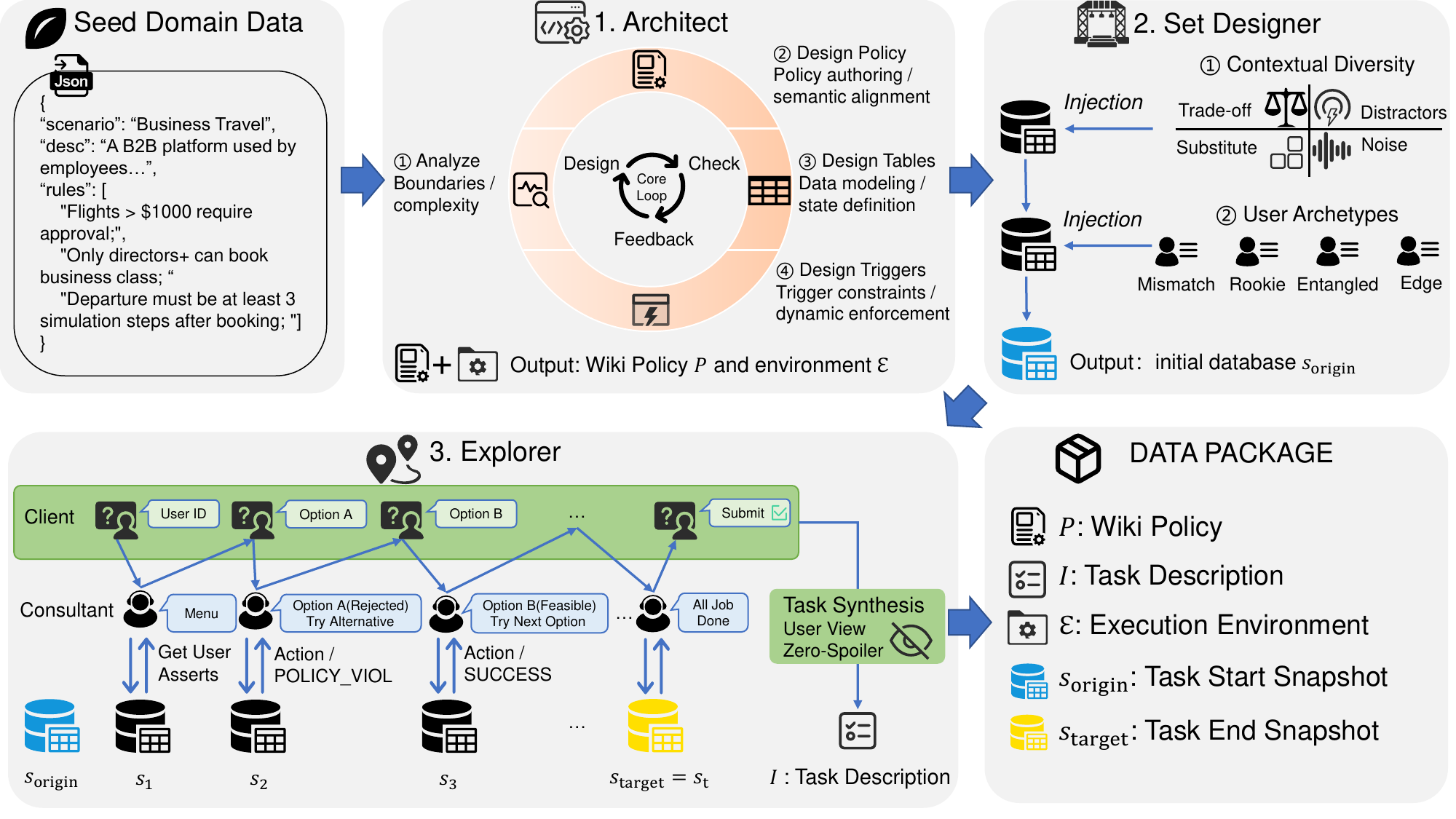}
  \caption{
\textbf{LOGIGEN} synthesizes verifiable agentic tasks via a \textbf{Triple-Agent Orchestration}: (\textbf{1}) the \textit{Architect} expands seed domain knowledge into a Wiki Policy and compiles it into a Hard-Compiled Policy Environment; (\textbf{2}) the \textit{Set Designer} seeds boundary-adjacent initial states ($N{-}1$) to maximize logical friction; and (\textbf{3}) the \textit{Explorer} performs goal-conditioned exploration to discover executable multi-turn episodes, producing a spoiler-free task description and a deterministic target database snapshot for state-based verification.
  }
  \label{fig:pipeline}
\end{figure}

\section{Method: LOGIGEN}
\label{sec:method}
\subsection{Problem Setup and Design Goals}
\label{sec:problem_setup}
To bridge the gap between the rigorous evaluation standards of $\tau$-Bench and the scarcity of high-quality training data, we introduce LOGIGEN, a logic-driven synthesis framework. LOGIGEN targets the state-transition objective: the framework is engineered to generate trajectories that successfully transform an initial database state $s_{\text{origin}}$ into a ground-truth target state $s_{\text{target}}$ while strictly adhering to complex wiki policy.

LOGIGEN operates through a hard-compiled policy environment and logic-driven deductive synthesis. Specifically, it (i) compiles natural-language wiki policy into physically enforced database constraints, ensuring that every action interacts with a rigorous logic engine; and (ii) performs deductive synthesis to synthesize tasks whose successful completion is strictly verifiable through deterministic State-Diff checks.

\textbf{Output Package.}
A single synthesized task instance is a self-contained package:
\begin{equation}
\mathcal{D}_i = \langle \mathcal{P}, \mathcal{I}, \mathcal{E}, s_{\text{origin}}, s_{\text{target}} \rangle,
\end{equation}
and the full dataset is $D=\{\mathcal{D}_i\}_{i=1}^N$.
Here, $\mathcal{P}$ is the \textbf{Wiki Policy} (natural-language business rules), $\mathcal{I}$ is a structured \textbf{task description} driving a user simulator.
The environment $\mathcal{E}$ is a \textbf{hard-compiled policy environment}:
\begin{equation}
\mathcal{E} = \langle \Sigma, \mathcal{T}, \texttt{executor}\rangle,
\end{equation}
where $\Sigma$ is the \textbf{database schema} (tables and triggers), $\mathcal{T}$ is a set of \textbf{atomic CRUD tools} with typed JSON schema, and \texttt{executor} is the Python + SQLite runtime that enforces database constraints and executes tools. Crucially, $\mathcal{P}$ provides the semantic context necessary for high-level planning, while $\mathcal{E}$ enforces the hard constraints at execution time. Finally, $s_{\text{origin}}$ and $s_{\text{target}}$ are SQLite snapshots defining the state-transition objective: the agent must successfully transform the environment from $s_{\text{origin}}$ to $s_{\text{target}}$ under $\mathcal{P}$ and $\mathcal{E}$.

\subsection{System Overview: Triple-Agent Orchestration}
Figure~\ref{fig:pipeline} illustrates the Triple-Agent Orchestration framework, which factorizes the synthesis of complex agentic tasks into three interdependent stages: \textbf{Policy Compilation}, \textbf{Boundary State Initialization}, and \textbf{Deductive Task Synthesis}.

\begin{enumerate}
\item \textbf{The Architect}: Expands minimal seed domain knowledge into a Wiki Policy $\mathcal{P}$ and compiles it into a physically enforced, \textbf{Hard-Compiled Policy Environment} $\mathcal{E}$.
\item \textbf{The Set Designer}: Initializes the environment to a database state $s_{\text{origin}}$, using an \textbf{boundary-adjacent} seeding principle.
\item \textbf{The Explorer}: Performs an exploration to \emph{discover} an executable trajectory, producing both a spoiler-free task description $\mathcal{I}$ and the corresponding ground-truth target state $s_{\text{target}}$.
\end{enumerate}

\subsection{The Architect: Compiling Policy into Environment}
The \textit{Architect} agent acts as a compiler that transforms minimal \textbf{seed domain data} into an execution environment $\mathcal{E}=\langle \Sigma,\mathcal{T},\texttt{executor}\rangle$.

\subsubsection{Policy Compilation and Environment Construction}
To ensure logical consistency and complexity, we adopt a four-stage compilation pipeline:
\begin{enumerate}
    \item \textbf{Analyze (Complexity Planning):}
    The agent first drafts a blueprint that explicitly injects \textbf{complex decision logic}, such as multi-variable conditional branches, state-dependent constraints, role-based permissions, and irreversible state transitions. This design prevents the system from degenerating into a simple, shallow CRUD application, ensuring that it supports sophisticated business workflows.
    \item \textbf{Policy Codification (Wiki Policy):}
    The agent expands the seed domain data into a comprehensive Wiki Policy $\mathcal{P}$. This document specifies business rules, access privileges, transaction flows, as well as the required data structures, and relational constraints. It serves as the semantic source-of-truth that guides the subsequent design of database tables and trigger logic.
    \item \textbf{Static Database Tables:}
    The agent translates the structural definitions in $\mathcal{P}$ into a normalized relational schema, defining tables and static integrity constraints (e.g., keys and data types). This step establishes the \textbf{static database foundation} required to support the business entities and defines the valid space of system states.
    \item \textbf{Dynamic Database Triggers:}
    The agent compiles the operational rules from $\mathcal{P}$ into SQL triggers that enforce policy logic at the database level.
    \begin{itemize}
        \item \textbf{Deterministic interception:} \texttt{BEFORE INSERT/UPDATE} triggers validate preconditions and raise a structured \texttt{POLICY\_VIOLATION} error on invalid writes.
        \item \textbf{Policy-mandated side effects:} \texttt{AFTER} triggers implement necessary bookkeeping (e.g., audit records, automatic inventory deduction), ensuring policy enforcement is handled by the database engine rather than relying on agent adherence.
    \end{itemize}
\end{enumerate}

\textbf{Iterative Verification Loop.}
To ensure strict adherence to the Wiki Policy $\mathcal{P}$ and prevent implementation drift, we employ a rigorous \textbf{Check--Fix--Verify} cycle after every compilation stage. A dedicated \textit{Verifier Agent} conducts a dual-layer validation:
\begin{itemize}
\item \textbf{Semantic Consistency:} The agent validates each generated artifact against its upstream specification (Policy $\rightarrow$ Tables $\rightarrow$ Triggers) using a chain-of-thought checklist derived from $\mathcal{P}$. This ensures that no logical requirements are omitted or misinterpreted during translation.
\item \textbf{Physical Executability:} The agent enforces runtime correctness by executing the generated SQLite DDL. A stage is accepted only if the tables and triggers compile successfully and all declared relations are instantiated without errors.
\end{itemize}
If a discrepancy is detected in either layer, the Verifier generates a structured error report, prompting the Architect to initiate a corrective iteration.

\subsubsection{Atomic Tool Generation with Logic-Exposed Interfaces}

Based on the schema $\Sigma$, we construct the agent-facing interface $\mathcal{T}$. We intentionally utilize atomic data manipulation tools (specifically per-table \texttt{insert}, \texttt{query}, and \texttt{update}) to preserve maximum planning freedom; the \texttt{delete} operation is excluded to encourage the use of soft-deletes via status fields.

\textbf{Standardized Error Feedback.}
To facilitate reliable self-correction, raw SQLite exceptions are intercepted and translated into a structured error payload:
\begin{equation}
\texttt{error}=\{\texttt{code},\texttt{message},\texttt{violated\_rule},\texttt{hint}\},
\end{equation}
ensuring that the agent receives actionable feedback instead of cryptic system errors.

\textbf{Logic-Aware Tool Descriptions.}
To mitigate the ``black-box'' effect of database triggers, we explicitly embed trigger logic within the tool schema:
\begin{itemize}
\item \textbf{Preconditions:} Conditions enforced by \texttt{BEFORE} triggers are documented as explicit preconditions or warnings.
\item \textbf{Side Effects:} Outcomes guaranteed by \texttt{AFTER} triggers are documented to prevent redundant writes or unintended state mutations.
\end{itemize}
This interface supports \textbf{logic-aware planning}, enabling agents to reason about constraints and effects without accessing privileged implementation details.

\subsection{The Set Designer: Critical State Seeding}
\label{sec:set_designer}
To ground the environment in a challenging context, the \textbf{Set Designer} initializes the database state $s_{\text{origin}}$ with structured diversity. The objective is to maximize \textbf{decision density} from the very first step, avoiding the introduction of random noise.

\subsubsection{Boundary-Adjacent Initialization}
We operationalize a \textbf{boundary-adjacent seeding} principle to maximize logical friction from the very first step. For constraints with numerical thresholds (e.g., a maximum capacity $N$), we initialize the database at the critical boundary $N-1$. For discrete or multi-variable states, we identify states that lie exactly one valid action away from triggering a rule violation. By initializing the environment at these critical junctures, we ensure the agent is immediately immersed in a high-stakes decision context, forcing it to \textbf{deduce the correct path under tight constraints} rather than merely executing linear commands.

\subsubsection{Two-Phase Construction}
Initialization proceeds in two coordinated phases to build a robust context:
\begin{enumerate}
\item \textbf{Resource Injection (Contextual Diversity):}
We populate resources employing four distinct strategies to induce environmental complexity:
\begin{itemize}
\item \textbf{Trade-offs:} Resources with conflicting attributes forcing prioritization.
\item \textbf{Distractors:} Plausible but incorrect options that mislead simple heuristics.
\item \textbf{Substitutes:} Functionally equivalent alternatives that require disambiguation.
\item \textbf{Noise:} High-cardinality irrelevant data that increases search difficulty.
\end{itemize}

\item \textbf{Cast Assembly (User Archetypes):}
We instantiate diverse user profiles covering four key archetypes to ensure broad scenario coverage:
\begin{itemize}
\item \textbf{Mismatch:} Users whose attributes conflict with resource requirements.
\item \textbf{Entangled:} Users embedded in complex multi-step dependency chains.
\item \textbf{Rookie:} Profiles with limited context, information, or privileges.
\item \textbf{Edge:} Users positioned precisely at constraint boundaries.
\end{itemize}
\end{enumerate}

This strategy is designed to complement the upstream Architect phase, relying on the policy's inclusion of \emph{reachable boundaries} (e.g., small integer thresholds and explicit state variables) to ensure that these critical, high-conflict states can be reliably constructed.

\subsection{The Explorer: Collaborative Causal Discovery}
The Explorer discovers executable multi-turn episodes via deductive exploration within the hard-compiled environment. Given $(\mathcal{P}, \mathcal{E}, s_{\text{origin}})$, it generates an episode $\tau$, a physically realized ground truth target database snapshot $s_{\text{target}}$, and a spoiler-free task description $\mathcal{I}$ derived from the user perspective.

\subsubsection{Enforcing Physical Causality}
We treat database state transitions as immutable physical facts. Each tool-induced write is executed by the database engine under the submerged constraints (tables, keys, and triggers) defined in $\mathcal{E}$.
An episode induces a state sequence  $s_0{=}s_{\text{origin}}, \dots, s_t$, and we define $s_{\text{target}}{=}s_t$ as the ground-truth database snapshot. Crucially, because $s_{\text{target}}$ is obtained via execution rather than prediction, it is guaranteed to be free of ground-truth drift.

\subsubsection{Dual-Role Collaborative Exploration}
To eliminate pre-scripted narrative bias, we decouple the ``goal owner'' from the ``policy executor'' using a Client--Consultant dynamic.

\textbf{The Consultant} (Policy-Aware) is state-grounded: it inspects $s_t$ via read tools and consults $\mathcal{P}$ to propose a feasible option menu set $\mathcal{O}(s_t)$.
\textbf{The Client} (Goal-Oriented) focuses on intent: it selects a user-level goal $g \in \mathcal{O}(s_0)$ and issues progressive, often imperfect requests (e.g., ambiguities, mind changes, incremental demands), relying on the Consultant for execution.
 
The exploration follows a structured loop:
\begin{enumerate}
    \item \textbf{Menu \& Goal Selection:} The \textbf{Consultant} proposes $\mathcal{O}(s_t)$ grounded in the current state, and the \textbf{Client} commits to a goal $g \in \mathcal{O}(s_t)$;
    \item \textbf{Preference Queries:} The \textbf{Client} asks attribute-level questions (e.g., ``cheapest''); the \textbf{Consultant} grounds comparisons via additional reads.
    \item \textbf{Parameter Resolution:} The \textbf{Client} makes descriptive choices (concepts, not IDs), and the \textbf{Consultant} resolves concrete identifiers through state queries.
    \item \textbf{Operation \& Negotiation:} The \textbf{Consultant} attempts goal-relevant operations (reads and writes). On \texttt{SUCCESS}, it confirms the operation, summarizes the outcome (including any trigger-induced side effects), and proposes a revised set of feasible options for the next step. On \texttt{POLICY\_VIOLATION}, it interprets the error using evidence from the current state and proposes viable alternatives, creating the recurrent loop: \textbf{Action $\rightarrow$ Trigger Feedback $\rightarrow$ Correction/Alternative}.
\end{enumerate}

Episodes are terminated early if they irreversibly diverge from the goal $g$ (e.g., repeated hard violations without new information).
 
\subsubsection{User-View Task Projection}
The raw episode contains privileged solution traces (tool calls, internal keys, and reasoning). We apply a strict \textbf{user-view projection} to generate a spoiler-free, goal-oriented task description:

\begin{itemize}
\item \textbf{One-Sided Translation (Goal-First, Procedure-Hidden):} We retain only public dialogue and user-visible facts to compile the task description $\mathcal{I}$. It anchors on the user's objectives and constraints (\emph{what/why}) while explicitly removing procedural cues (\emph{how}), such as tool names, table schemas, and internal identifiers.
\item \textbf{Two-View protocol:} the trainee agent receives $(\mathcal{P}, \mathcal{I}, \mathcal{E}, s_{\text{origin}})$, while $s_{\text{target}}$ is reserved strictly for evaluation and reward computation.
\end{itemize}

\subsection{Dataset Characteristics}
\label{sec:dataset_characteristics}
We instantiate the LOGIGEN pipeline across a diverse taxonomy of \textbf{8 core domains} and multiple sub-scenarios (see Table~\ref{tab:domains}). This process yields a large-scale, logic-dense dataset comprising \textbf{over 20,000} unique, verifiable tasks. Furthermore, to construct the training corpus for Supervised Fine-Tuning (SFT), we synthesized a collection of \textbf{verified SFT trajectories} through closed-loop execution within the generated environments. We utilized \textbf{DeepSeek-V3.2} to power the User Simulator, which generated natural language intents driven by the task descriptions $I$, while a Teacher Agent based on \textbf{DeepSeek-V3.2-Thinking} was tasked with tool execution and planning. Crucially, we applied rigorous filtering based on deterministic state checks: only trajectories that successfully transformed the initial state $s_\text{origin}$ into the target state $s_\text{target}$ were retained. To provide a concrete illustration of the task packages generated by our Triple-Agent Orchestration, we present a detailed case study in Appendix~\ref{appendix:case_study}. Further details on data generation costs and the User Simulator prompt are provided in Appendix~\ref{appendix:dataset_construction}.

\subsubsection{Scale and Environment Diversity}

The Architect compiles natural language policies into \textbf{over 2,000} distinct hard-compiled policy environments $\mathcal{E}$, each characterized by unique policy, database schema, and toolsets. By applying the Set Designer’s boundary-adjacent seeding and the Explorer's collaborative causal discovery, we generate approximately 10 tasks per environment on average. Each task $\mathcal{D}_i$ serves as a self-contained package, enabling both state-based training and deterministic evaluation via the $(s_{\text{origin}}, s_{\text{target}})$ pair.

\subsubsection{Complexity Profile and Reasoning Depth}

To characterize the reasoning requirements of the generated tasks, we apply a multi-dimensional tagging rubric. Figure~\ref{fig:task_statistic}(a), the dataset is predominantly composed of high-order agentic challenges:

\begin{enumerate}
    \item \textbf{Logic-Intensive Reasoning:} \textsc{Conditional-Logic} (7,232) and \textsc{Multi-Step} (6,514) tasks constitute the majority, requiring agents to navigate complex branching structures and long-horizon interactions.
    \item \textbf{Robustness \& Fallback:} A significant portion involves \textsc{Waterfall} (3,203), \textsc{Adversarial} (2,384), and \textsc{Hidden-Constraint} (1,709) scenarios, which compel the agent to handle tool failures, policy violations, or implicit user needs.
    \item \textbf{State Inference:} \textsc{State-Based} (3,406) and \textsc{Reasoning} (2,333) tags highlight tasks where agents must derive plans from goal states rather than following explicit procedural instructions.
\end{enumerate}

\subsubsection{Difficulty Stratification}

We categorize tasks into three difficulty tiers based on the reasoning depth and interaction friction required to reach $s_{\text{target}}$:

\begin{enumerate}
    \item \textbf{Level 1 (Simple):} Tasks involving \textbf{linear execution} where the agent follows direct commands with explicit parameters (e.g., specific IDs or dates) and encounters minimal logical obstacles.
    \item \textbf{Level 2 (Intermediate):} Tasks requiring \textbf{conditional decision-making}, such as attribute-based selection (e.g., "find the cheapest option") or resolving goals through multi-turn clarifications.
    \item \textbf{Level 3 (Advanced):} Tasks designed with \textbf{high-order logical friction}, including:
    \begin{itemize}
        \item \textbf{State-based Reasoning:} Deducing specific tool-calls from a desired end-state rather than procedural instructions.
        \item \textbf{Adversarial Interactions:} Handling user misconceptions, false beliefs, or stubbornness.
        \item \textbf{Waterfall Logic:} Navigating multiple layers of fallback strategies (e.g., \textit{if A fails, try B; if B fails, negotiate C}).
    \end{itemize}
\end{enumerate}

\begin{figure*}[h]
    \centering
    \begin{subfigure}[b]{0.7\textwidth}
        \centering
        \includegraphics[width=\textwidth]{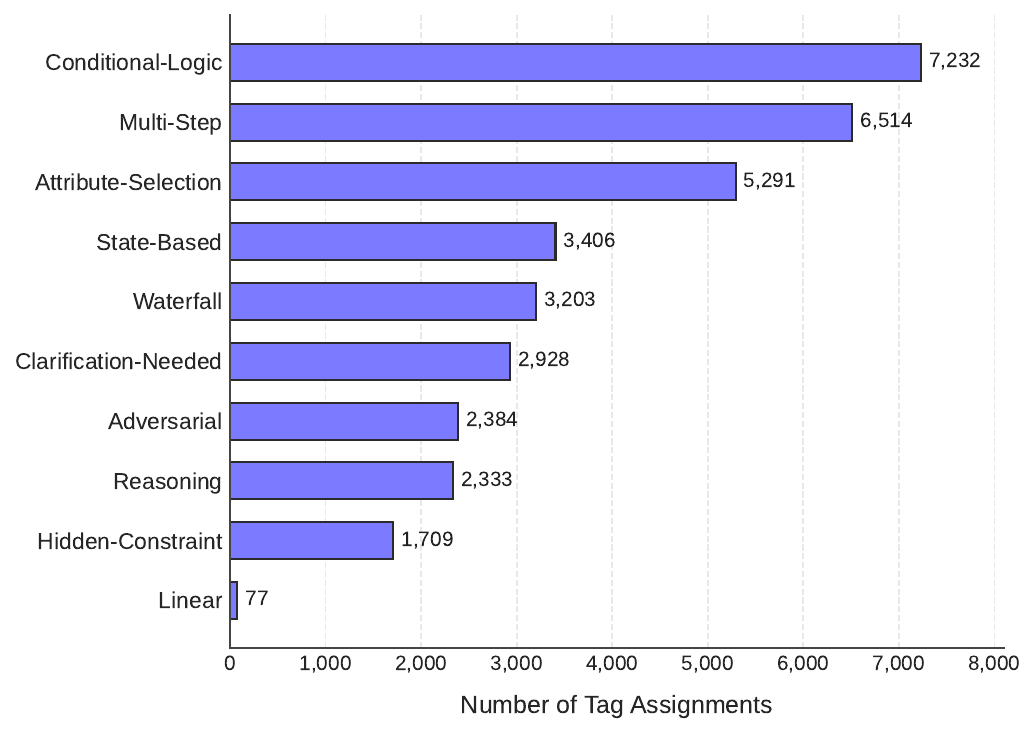}
        \caption{Distribution of complexity tags in the generated task.}
        \label{fig:perf_comparisons}
    \end{subfigure}
    \hfill
    \begin{subfigure}[b]{0.29\textwidth}
        \centering
        \includegraphics[width=\textwidth]{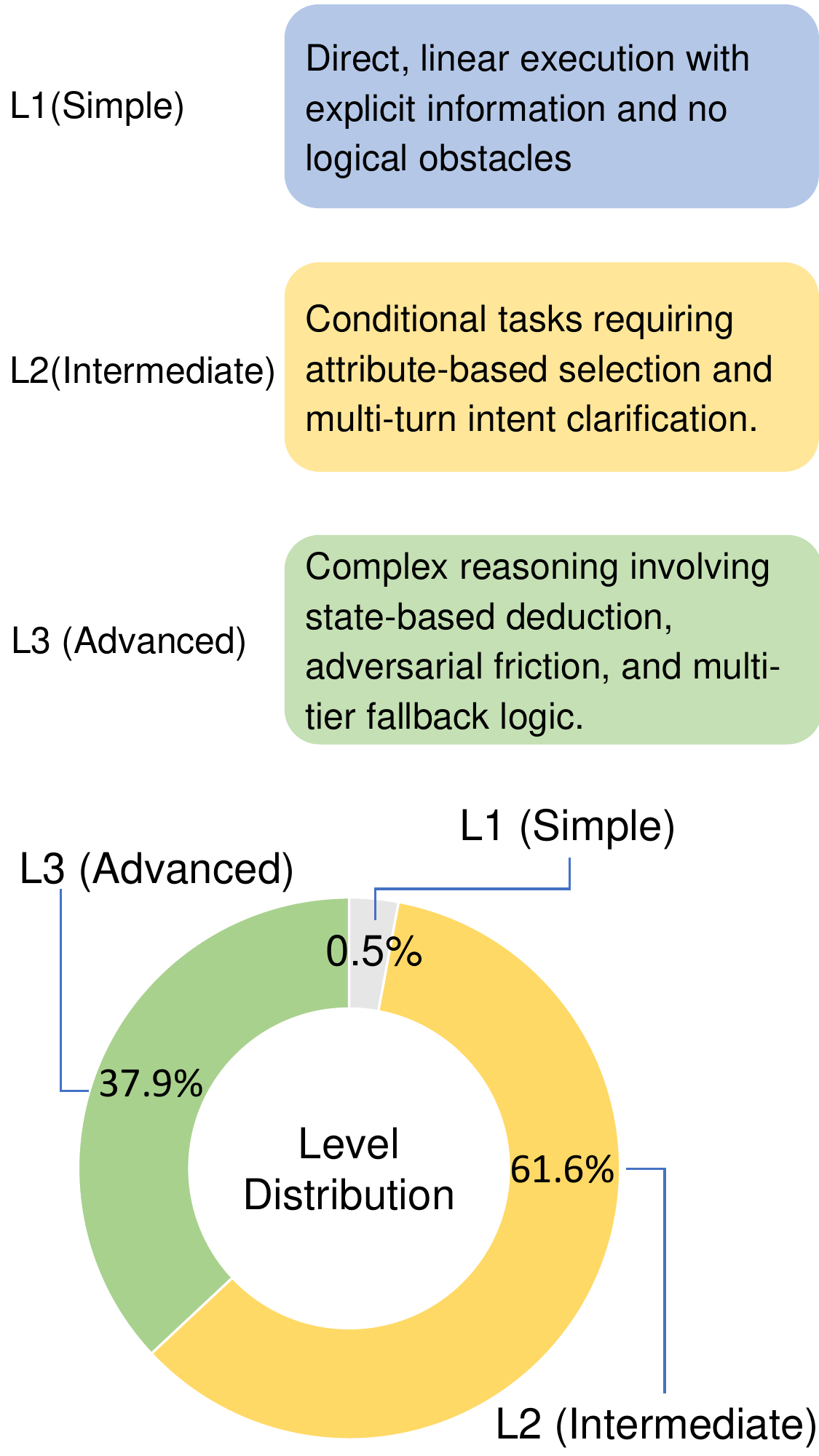}
        \caption{Complexity Level Distribution}
        \label{fig:incre_gains}
    \end{subfigure}
    \caption{
    \textbf{Task complexity profile of LOGIGEN.}
    \textbf{(a)} Distribution of the top-10 complexity tags assigned to generated tasks, showing that most samples involve conditional logic, multi-step interaction, and attribute-based selection, with a substantial portion requiring state-based reasoning and fallback (waterfall) behaviors.
    \textbf{(b)} Difficulty-level distribution, where L1 (Simple) tasks are rare (51), while the dataset is dominated by L2 (Intermediate; 6,161) and L3 (Advanced; 3,788) tasks, indicating a strong bias toward non-trivial, policy-constrained problem solving.
    }
    \label{fig:task_statistic}
\end{figure*}

As illustrated in Figure~\ref{fig:task_statistic}(b), the dataset is intentionally skewed toward complexity: L2 (6,161) and L3 (3,788) tasks represent over 99\% of the total, while trivial L1 cases (51) are negligible. This distribution underscores LOGIGEN’s capacity to synthesize non-linear, policy-constrained problems that push the limits of current LLM-based agents.

\begin{table}[htbp]
    \centering
    \caption{Taxonomy of Core Domains and Scenarios}
    \label{tab:domains}
    \begin{tabularx}{\linewidth}{>{\bfseries}l X}
        \toprule
        Core Domain & Sub-domains \& Scenarios \\
        \midrule
        Travel \& Hospitality & Travel Agent, Hotel Concierge, Corporate Travel Booking, \newline Hospitality \& Travel Services, Event Ticketing \\
        \addlinespace
        Retail \& Supply Chain & Order Assistant, After-Sales Support, Shopping Assistant, \newline Membership Support, Logistics Support, Supply Chain \& Logistics \\
        \addlinespace
        Healthcare \& Life Sci. & Medical Receptionist, Medical Triage \& Booking, \newline Healthcare \& Life Sciences Services \\
        \addlinespace
        Finance \& Insurance & Bank Teller Bot, Insurance Claims, Insurance Services, \newline Finance \& Account Steward \\
        \addlinespace
        Education \& Knowledge & University Registrar, Course Enrollment, Librarian, Skill Development \\
        \addlinespace
        Enterprise Ops \& HR & HR Assistant, HR Team Management, Public \& Admin Clerk, \newline Collaboration Creation, Compliance Process, Planning \& Strategy \\
        \addlinespace
        Real Estate \& Lifestyle & Property Manager, Real Estate \& Property Management, \newline Life Services Assistant, Dining Reservation \\
        \addlinespace
        Technology \& Media & Telecom Support, Diagnosis Troubleshooting, \newline Gaming Community Platforms, Media Content Platforms \\
        \bottomrule
    \end{tabularx}
\end{table}

\begin{figure}[t]
  \centering
  \includegraphics[width=\linewidth]{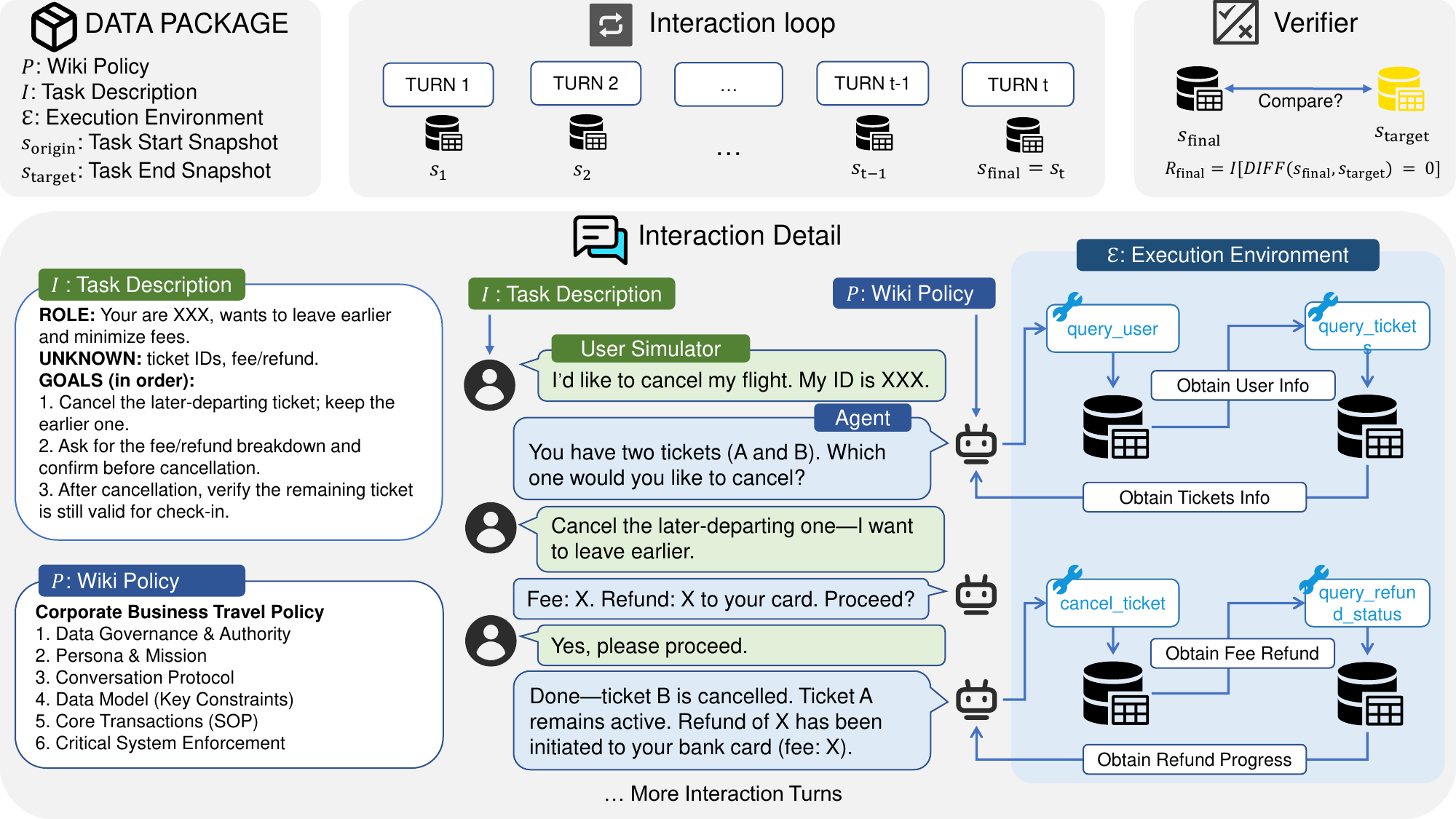} 
  \caption{
    \textbf{Training and Evaluation Protocol of LOGIGEN.} The protocol operates on a \textbf{Data Package} containing the Wiki Policy ($\mathcal{P}$), Task Description ($\mathcal{I}$), and paired database snapshots. Within the \textbf{Interaction Loop}, a \textbf{User Simulator} (driven by $\mathcal{I}$) interacts with the \textbf{Agent} (governed by $\mathcal{P}$ and tools in $\mathcal{E}$). Each tool invocation induces a deterministic mutation of the database state ($s_\text{origin} \rightarrow s_\text{1} \rightarrow s_\text{2} \rightarrow \dots \rightarrow s_\text{final}$). The \textbf{Verifier} computes a binary success metric $R_{final}$ by performing a canonicalized \textit{State-Diff} between the resulting state $s_\text{final}$ and the ground-truth target $s_\text{target}$.
  }
  \label{fig:interact_protocol}
\end{figure}

\section{Training and Evaluation Protocol}

Each sample in the LOGIGEN dataset serves as a fully executable, self-contained specification. This protocol enables practitioners to train and evaluate agents within a closed-loop interaction between a \textbf{User Simulator} and a \textbf{Tool-Calling Agent}, where correctness is judged solely by the resulting state against the ground-truth target.

\subsection{Data Package and Rollout Interface}

As illustrated in Figure\ref{fig:interact_protocol}, a LOGIGEN sample instantiates an interactive task via the data package $\mathcal{D}_i = \langle \mathcal{P}, \mathcal{I}, \mathcal{E}, s_{\text{origin}}, s_{\text{target}} \rangle$, formally defined in Section~\ref{sec:problem_setup}. This package encapsulates the entire lifecycle of a rollout, ranging from initial environment seeding to terminal state verification.

\paragraph{Initialization.}
The environment is instantiated by loading $s_{\text{origin}}$ into a sandboxed database engine. The agent is provided with $\mathcal{P}$ and tool definitions $\mathcal{T}$, while the User Simulator is initialized with $\mathcal{I}$.

\paragraph{Interaction Loop.}
At each turn $t$, the User Simulator generates a natural-language utterance $u_t$ expressing high-level intent, compelling the agent to perform task decomposition. The agent responds with either (a) a tool call $(\mathbf{t}, \mathbf{a})$ where $\mathbf{t} \in \mathcal{T}$, or (b) a natural-language response. Critically, tool execution deterministically mutates the database state ($s_\text{t} \rightarrow s_\text{t+1}$) and returns a structured result, which is either a success payload or a structured error object derived from policy violations.

\paragraph{Loop Termination.}
The episode terminates when (i) the User Simulator emits a stop signal (e.g., \#\#\#STOP\#\#\#) indicating goal satisfaction under $\mathcal{I}$, (ii) an \emph{irrecoverable deviation} from the task specification $\mathcal{I}$ is detected, or (iii) a maximum turn budget is exhausted.

\subsection{Deterministic and Dense Reward Mechanism}
\label{sec:deterministic}
LOGIGEN enables \textbf{deterministic state-transition evaluation} that is objective and path-independent. Unlike conventional benchmarks that rely on fragile LLM-based grading or syntactic matching, we define success through precise database state equivalence.

Let $s_t$ denote the database state at turn $t$, and $s_\text{{final}}$ denote the terminal state. We define a distance function $\mathrm{DIFF}(s_\text{a}, s_\text{b})$ as the cardinality of the symmetric difference between the canonicalized relational sets of two snapshots:

$$ \mathrm{DIFF}(s_a, s_b) = \sum_{T \in \mathcal{V}} | \text{rows}(T, s_\text{a}) \triangle \text{rows}(T, s_\text{b}) | $$

where $\mathcal{V}$ is the set of database tables and $\triangle$ is the symmetric set difference. \textbf{Crucially, to ensure semantic equivalence, we exclude technical database keys (e.g., auto-incrementing IDs and timestamps) from the comparison. This ensures that $\mathrm{DIFF}(\cdot)$ measures logical business outcomes rather than procedural artifacts.}

\paragraph{Binary success.}
The agent achieves success ($R_\text{{final}} = 1$) if and only if the final database state is semantically indistinguishable from the target state:
$$ R_\text{{final}} = I \big[ \mathrm{DIFF}(s_\text{{final}}, s_{\text{target}}) = 0 \big] $$
\paragraph{State Proximity and Dense Rewards.}
While $R_\text{{final}}$ provides the ultimate standard for evaluation, the database-backed nature of LOGIGEN allows us to derive a granular, continuous measure of progress, addressing the sparse reward problem in long-horizon RL.

Let $\Delta_0 = \mathrm{DIFF}(s_{\text{origin}}, s_{\text{target}})$ represent the total logical distance of the task. We define a normalized \textbf{State Proximity Score} $P_t \in [0, 1]$ at turn $t$ as:

$$P_t = 1 - \frac{\min\left(\mathrm{DIFF}(s_\text{t}, s_{\text{target}}), \Delta_0\right)}{\Delta_0 + \epsilon}$$

Here, $P_t = 1$ indicates that the agent has successfully reached the goal state.

Based on this proximity, we derive a \textbf{Dense Incremental Reward} $r_t$ for reinforcement learning. To explicitly penalize policy violations caught by the hard-compiled constraints (which trigger database rollbacks and leave the state unchanged), we formulate the reward as:

\begin{equation}
r_t = 
\begin{cases} 
P_t - P_{t-1}, & \text{if execution succeeds} \\ 
-\lambda_{\text{err}}, & \text{if policy violation occurs} 
\end{cases}
\end{equation}

where $\lambda_{\text{err}} > 0$ is a penalty coefficient. Positive $r_t$ rewards actions that bring the state closer to $s_{\text{target}}$ (e.g., fulfilling a sub-goal), while negative $r_t$ penalizes policy violations or actions that increase the distance (e.g., executing an irrelevant operation that disrupts the current progress). This dense supervision provides fine-grained feedback essential for credit assignment in multi-turn tasks.

\subsection{Turn-aware GRPO for Long-Horizon Optimization}
\label{sec:ta_grpo}
To optimize the agent policy using the dense rewards defined above, we adopt Group Relative Policy Optimization (GRPO)~\citep{grpo} and introduce a refinement mechanism to address the credit assignment challenges specific to multi-turn agentic workflows.

\paragraph{Limitations of Vanilla GRPO.}
GRPO computes advantage functions through intra-group relative rewards, eliminating the need for complex Value Function approximations. In standard applications, GRPO treats the entire trajectory as a unit: given a group of $G$ trajectories $\{y_1, \dots, y_G\}$, it computes a trajectory-level advantage $A_i$ by normalizing the final rewards $\{R_i\}_{i=1}^G$:

$$A_i = \frac{R_i - \text{mean}(\{R_i\})}{\text{std}(\{R_i\})}$$

In long-horizon tasks, however, assigning a uniform $A_i$ to every turn within a trajectory creates a severe bottleneck. Even in a successful trajectory, intermediate steps may contain erroneous actions (e.g., making a mistake later corrected by another tool). Assigning a positive $A_i$ to these flawed steps misleads the model, preventing it from learning to avoid specific local errors.

\paragraph{Turn-aware GRPO (TA-GRPO).} To provide more granular gradient guidance, we refine the advantage function at the turn level using the dense reward signal $r_t$ defined in Section~\ref{sec:deterministic}.

We define the refined turn-level advantage $A_{it}$ for turn $t$ in trajectory $i$ as:

$$A_{it} = A_i + \Delta A_{it}$$

Specifically, we implement an asymmetric refinement strategy based on the incremental reward $r_t$:

$$
\Delta A_{it} = 
\begin{cases} 
r_t, & \text{if } r_t < 0 \\
0, & \text{if } r_t \ge 0
\end{cases}
$$

\begin{itemize}
    \item \textit{Negative Progress ($r_t < 0$):} If an action degrades the task progress (e.g., accidental deletion of data or a policy violation), we add the negative $r_t$ to the advantage. This explicitly penalizes the step, forcing the model to suppress harmful actions even if they occur in a trajectory that eventually succeeds.
    \item \textit{Non-negative Progress ($r_t \ge 0$):} While a positive $r_t$ indicates progress, it does not guarantee the action was optimal. To prevent the model from becoming overconfident in sub-optimal paths and falling into local optima, we do not provide additional positive bonuses. We preserve the group relative advantage $A_i$ to maintain optimization stability.
\end{itemize}

\paragraph{Optimization Objective.} The final optimization objective for TA-GRPO is formulated as:

\begin{equation}
\begin{aligned}
J_{\text{TA-GRPO}}(\theta) = E_{q \sim D, \{y_i\} \sim \pi_{\text{old}}} \Bigg[ &\frac{1}{G} \sum_{i=1}^{G} \sum_{t=1}^{T} \Big( \\
&\min \left( \frac{\pi_{\theta}(y_{it}|q)}{\pi_{\text{old}}(y_{it}|q)} A_{it}, \text{clip}\left(\frac{\pi_{\theta}(y_{it}|q)}{\pi_{\text{old}}(y_{it}|q)}, 1 - \epsilon, 1 + \epsilon\right) A_{it} \right) \\
&- \beta D_{KL}(\pi_{\theta} \| \pi_{\text{ref}}) \Big) \Bigg]
\end{aligned}
\end{equation}

By integrating step-level progress feedback, TA-GRPO preserves the stability and efficiency of GRPO while significantly enhancing the model's ability to discern and rectify local errors within complex interaction trajectories.

\subsection{Supported Training Paradigms}

The data packages and protocol described above support multiple training paradigms, directly addressing the bottlenecks identified in Section~\ref{sec:introduction}:

\begin{itemize}
    \item \textbf{Benchmark Evaluation.} Practitioners can run one or multiple rollouts per task and report success rates (e.g., Pass \^\ k) using the binary metric $R_\text{final}$. 
    \item \textbf{Verified SFT.} We ensure logical consistency by executing trajectories from a teacher model and rigorously filtering them. Only trajectories achieving $R_\text{final}=1$ are retained, which guarantees that the expert demonstrations strictly satisfy the hard-compiled policy and physical state transitions.
    \item \textbf{State-Based RL.} The deterministic state reward signal enables precise, automated feedback for optimization. Using $R_\text{final}$ as a sparse terminal reward or $\{r_t\}$ as dense intermediate rewards, agents can be trained to master complex, long-horizon tasks.
\end{itemize}

\begin{table}[htbp]
\centering
\small
\renewcommand{\arraystretch}{1.2} 
\setlength{\tabcolsep}{4pt}       

\caption{Main results on $\tau^{2}$-Bench.}
\label{tab:tau2_results}

\begin{tabularx}{\linewidth}{X l c c c c}
\toprule
\textbf{Model} & \textbf{User Simulator} & \multicolumn{4}{c}{\textbf{$\tau^{2}$-Bench}} \\
\cmidrule(l){3-6}
& & Retail & Airline & Telecom & Overall \\
\midrule 

\rowcolor{gray!10} 
\multicolumn{6}{l}{\textbf{Closed-Source Large Language Models}} \\
\midrule 

Gemini-3-pro       & Gemini-3-pro       & 85.3 & 73.0 & 98.0 & 85.4 \\
Claude-Sonnet-4.5  & -                  & 86.2 & 70.0 & 98.0 & 84.7 \\
GPT-5              & GPT-4.1-2025-04-14 & 81.6 & 62.5 & 95.8 & 80.0 \\
Qwen3-Max-Thinking & GPT-4.1-2025-04-14 & 79.4 & 69.0 & 98.2 & 82.2 \\
Qwen3-Max          & GPT-4.1-2025-04-14 & 72.2 & 59.5 & 84.2 & 72.0 \\
\addlinespace 

\midrule
\rowcolor{gray!10}
\multicolumn{6}{l}{\textbf{Open-Source Large Language Models}} \\
\midrule

GPT-OSS-120B-A5B            & -                  & 57.0 & 38.0 & 45.6 & 46.9 \\
DeepSeek-V3.2-Thinking      & DeepSeek-V3.2      & 81.1 & 63.8 & 96.2 & 80.4 \\
DeepSeek-V3.1-Terminus-Thinking    & DeepSeek-V3.2      & 65.4 & 44.0 & 23.7 & 44.4 \\
MiniMax-M2                  & -                  & -    & -    & 87.0 & 77.2 \\
Kimi-K2-Thinking            & -                  & 70.6 & 56.5 & 65.8 & 64.3 \\
LongCat-Flash-Thinking      & GPT-4.1-2025-04-14 & 71.5 & 67.5 & 83.1 & 74.0 \\
LongCat-Flash-Thinking-2601 & GPT-4.1-2025-04-14 & 88.6 & 76.5 & 99.3 & 88.2 \\
GLM-4.7-Thinking            & -                  & -    & -    & -    & 87.4 \\
Qwen3-235B-A22B-Thinking-2507 & -                & 71.9 & 58.0 & 45.6 & 58.7 \\
Qwen3-8B ‡                  & DeepSeek-V3.2      & 38.6 & 30.5 & 23.3 & 30.8 \\
Qwen3-32B ‡                 & DeepSeek-V3.2      & 53.3 & 42.0 & 26.9 & 40.7 \\

\addlinespace

\midrule
\rowcolor{gray!10}
\multicolumn{6}{l}{\textbf{Alternative Agent Training Frameworks}} \\
\midrule

TOUCAN-7B           & GPT-4o             & 22.8 & 20.0 & 10.5 & 17.7 \\
TOUCAN-32B          & GPT-4o             & 52.6 & 22.0 & 20.2 & 31.6 \\
AgentScaler-4B      & -                  & 62.3 & 56.0 & 48.2 & 55.5 \\
AgentScaler-8B      & -                  & 58.8 & 44.0 & 45.4 & 49.4 \\
AgentScaler-30B-A3B & -                  & 70.2 & 60.0 & 55.3 & 61.8 \\
MUA-RL-8B           & GPT-4.1-2025-04-14 & 49.8 & 19.0 & 21.8 & 30.2 \\
MUA-RL-32B          & GPT-4.1-2025-04-14 & 67.3 & 45.4 & 28.3 & 47.0 \\
LongCat-GEM-8B      & GPT-4.1-2025-04-14 & 44.5 & 22.0 & -    & - \\
LongCat-GEM-32B     & GPT-4.1-2025-04-14 & 55.5 & 35.5 & -    & - \\
EnvScaler-4B        & GPT-4.1-2025-04-14 & 48.1 & 34.0 & -    & - \\
EnvScaler-8B        & GPT-4.1-2025-04-14 & 53.6 & 36.0 & -    & - \\
Nemotron-3-Nano-30B-A3B & -              & 56.9 & 48.0 & 42.2 & 49.0 \\
\addlinespace

\midrule
\rowcolor{gray!10}
\multicolumn{6}{l}{\textbf{Our Work}} \\
\midrule

\textbf{LOGIGEN-8B(SFT)}  & DeepSeek-V3.2 & 74.1 & 61.5 & 80.7 & 72.1 \\
\textbf{LOGIGEN-8B(RL)}   & DeepSeek-V3.2 & 79.5 & 54.7 & 81.3 & 71.8 \\
\textbf{LOGIGEN-32B(SFT)} & DeepSeek-V3.2 & 82.0 & 64.0 & 86.6 & 77.5 \\
\textbf{LOGIGEN-32B(RL)}  & DeepSeek-V3.2 & \textbf{85.1} & \textbf{64.7} & \textbf{88.6} & \textbf{79.5} \\
\bottomrule
\end{tabularx}
\begin{tablenotes}[flushleft]
    \scriptsize
    \item[‡] ‡: Results obtained from our internal evaluation using the same protocol.
\end{tablenotes}
\end{table}

\section{Experiments}
\subsection{Setup}
\paragraph{Benchmarks.} We evaluate our models on $\tau^2$-Bench~\citep{tau2}, which extends the original $\tau$-Bench~\citep{tau1} by refining the Retail and Airline domains and introducing a complex, dual-control Telecom domain. Following standard protocols, we report \textbf{Pass \^\ 1} (averaged over four independent runs) to ensure robustness. We deliberately omit atomic function-calling benchmarks (e.g., BFCL~\citep{bfcl}, ACE-Bench~\citep{acebench}), as they focus on syntactic adherence rather than the causal reasoning required to achieve verifiable state transitions.

\paragraph{Baselines.} We compare LOGIGEN against three distinct categories of baselines to ensure a comprehensive evaluation:
\begin{itemize}
\item \textbf{Closed-Source Models:} Gemini-3-pro~\citep{Gemini3}, Claude-Sonnet-4.5~\citep{Claude-Opus-4.5}, GPT-5~\citep{gpt5} and Qwen3-Max~\citep{Qwen3-MAX}.
\item  \textbf{Open-Source Models:} GPT-OSS-120B-A5B~\citep{gpt-oss-120b}, DeepSeek-V3.1-Terminus-Thinking~\citep{DeepSeek-V31}, DeepSeek-V3.2-Thinking~\citep{DeepSeek-V31}, MiniMax-M2~\citep{MiniMax-M2}, KimiK2-Thinking~\citep{kimi-k2}, LongCat-Flash-Thinking~\citep{LongCat-Flash-Thinking}, LongCat-Flash-Thinking-2601~\citep{LongCat-Flash-Thinking-2601}, GLM-4.7-Thinking~\citep{GLM-4.7}, Qwen3-235B-A22B-Thinking-2507~\citep{qwen3}, Qwen3-8B~\citep{qwen3} and Qwen3-32B~\citep{qwen3}.
\item \textbf{Alternative Agent Training Frameworks:} TOUCAN~\citep{toucan}, AgentScaler~\citep{agentscaler}, MUA-RL~\citep{mua-rl}, LONG-CAT-GEM~\citep{longcat-gem}, EnvScaler~\citep{envscaler} and Nvidia-Nemotron-3-Nano-30b-A3B~\citep{nemotron3}.
\end{itemize}

\paragraph{Models.} The LOGIGEN series is built upon Qwen3~\citep{qwen3} base models of varying sizes. We train two primary variants to evaluate the impact of different training protocol: 
(1) \textbf{LOGIGEN-{\{8B,32B\}}+SFT}, derived via Verified Supervised Fine-Tuning on the full dataset; and 
(2) \textbf{LOGIGEN-{\{8B,32B\}}+RL}, developed through a two-stage pipeline comprising Cold-Start Fine-tuning followed by State-Based Reinforcement Learning.
During inference, all LOGIGEN models operate in a thinking-enabled mode, retaining reasoning traces across tool calls to ensure logical continuity, while resetting the context upon new user input.

\paragraph{Supervised Fine-Tuning.} We conduct full-parameter fine-tuning on the \textbf{20,000 verified trajectories} (described in Section~\ref{sec:dataset_characteristics}) using LLaMA-Factory~\citep{llamafactory}. The training spans two epochs with a constant learning rate of $5\times10^{-6}$ and a batch size of 64. We employ the Deepspeed ZeRO-3 strategy with BF16 precision, setting the maximum sequence length to 32,768 tokens and weight decay to 0.05. The Hermes template is utilized for tool-calling.

\paragraph{Cold-start Fine-tuning For RL.} We perform full-parameter fine-tuning on approximately $3,000$ synthesized trajectories, adhering to the same hyperparameter setup as the main SFT stage to ensure consistency.

\paragraph{Reinforcement Learning Training.} We implemented a multi-turn RL framework based on VolcEngine RL (VeRL)~\citep{verl}, integrated with a live database environment to validate tool invocations. We employ the Turn-aware GRPO (TA-GRPO) algorithm (Section~\ref{sec:ta_grpo}) for turn-level advantage refinement, setting the KL penalty coefficient to 0. The training configuration consists of 1,000 steps with a batch size of 64 and 8 rollouts, using a constant learning rate of $1 \times 10^{-6}$ and a maximum sequence length of 32,768 tokens. DeepSeek-V3.2 serves as the User Simulator (see Section~\ref{appendix:user_simulator} for the prompt), with a temperature of 1.0 applied to both the simulator and the agent. We cap interactions at 50 turns per task to ensure computational efficiency. Finally, a loss masking strategy is applied to tokens from tool results and user messages, allowing the model to focus on learning effective tool-use and communication patterns.

\subsection{Main Results}

Table~\ref{tab:tau2_results} presents the performance of various models across three real-world domains. The results validate the effectiveness of the LOGIGEN pipeline in synthesizing high-quality, logic-dense training data.

\paragraph{Narrowing the Gap between Open and Closed Models.} While closed-source LLMs like \texttt{Gemini-3-pro} (85.4\% Overall) and \texttt{Claude-Sonnet-4.5} (84.7\% Overall) currently represent the performance ceiling, the LOGIGEN series demonstrates that logic-dense synthetic trajectories can effectively bridge the competitive gap. Notably, \texttt{LOGIGEN-32B(RL)} achieves an overall score of \textbf{79.5\%}, outperforming the original \texttt{Qwen3-32B} base model by a remarkable absolute margin of 38.8\%. This performance places it in direct competition with state-of-the-art general-purpose models such as \texttt{GPT-5} (80.0\%) and \texttt{DeepSeek-V3.2-Thinking} (80.4\%). This suggests that agentic competence is not solely a function of model scale, but is heavily dependent on the logical density of the training data.

\paragraph{Superiority over Alternative Frameworks.} LOGIGEN consistently outperforms all baselines in the ``Alternative Agent Training Frameworks'' category. Specifically, \texttt{LOGIGEN-8B(RL)} (71.8\% Overall) significantly surpasses its direct competitors like \texttt{AgentScaler-8B} (49.4\%) and \texttt{MUA-RL-8B} (30.2\%). Remarkably, our 8B model even outperforms much larger alternative frameworks, such as \texttt{MUA-RL-32B} (47.0\%) and \texttt{AgentScaler-30B-A3B} (61.8\%). These results decisively validate the holistic design of the LOGIGEN architecture. By integrating hard-compiled policy enforcement, boundary-adjacent initialization, and deductive trajectory synthesis, our framework provides a rigorously grounded supervisory signal that is fundamentally superior to the ungrounded reverse-synthesis methods of prior work.

\paragraph{Exceptional Parameter Efficiency.} LOGIGEN exhibits exceptional parameter efficiency, outperforming general-purpose models within the same family that possess significantly larger parameter counts. For instance, \texttt{LOGIGEN-32B(RL)} (79.5\%) substantially surpasses \texttt{Qwen3-Max} (72.0\%), while even the compact \texttt{LOGIGEN-8B(RL)} (71.8\%) achieves performance on par with \texttt{Qwen3-Max}, demonstrating a remarkable efficiency gain. Furthermore, our 32B model delivers results competitive with state-of-the-art open-source models like \texttt{DeepSeek-V3.2-Thinking} (80.4\%). These findings suggest that general-purpose pre-training often yields suboptimal inductive biases for agentic tasks: while large models acquire broad linguistic capabilities, they struggle to efficiently internalize the rigid causality and state dependencies required for policy-governed environments. By training on logic-dense, verifiable trajectories, LOGIGEN models effectively specialize their parameter capacity towards rigorous logical deduction and constraint satisfaction, thereby avoiding the inefficient exploration patterns typical of larger, unguided models.

\paragraph{Impact of State-Based RL.} Comparing our SFT and RL variants, we observe that reinforcement learning provides crucial refinement, particularly for larger models. In the 32B parameter regime, RL fine-tuning boosts the overall success rate from 77.5\% to 79.5\%, with specific gains observed in Telecom and Airline domains. For the 8B model, RL optimization maintains competitive performance (71.8\%) while notably improving specific domains like Retail (+5.4\%). This improvement validates the critical role of deterministic state-based rewards. By optimizing the granular State-Diff ($r_t$), the agent receives precise, objective feedback for every action, enabling it to navigate complex policy constraints more effectively than supervised imitation alone.

\section{Analysis}

\subsection{Impact of Training Pipeline Stages}
\label{subsec:training_pipeline}

\definecolor{perfgreen}{HTML}{009966} 
\definecolor{perfred}{HTML}{CC0000}   

\newcommand{\up}[1]{\textcolor{perfgreen}{\small($\uparrow$#1)}}
\newcommand{\down}[1]{\textcolor{perfred}{\small($\downarrow$#1)}}

\begin{table}[htbp]
    \centering
    \caption{Incremental performance gains of training stages on $\tau^2$-Bench.}
    \small
    \setlength{\tabcolsep}{4pt}
    \renewcommand{\arraystretch}{1.2}
    \begin{tabular}{l c c c c}
    \toprule
    \multirow{2}{*}{\textbf{Model}} & \multicolumn{4}{c}{\textbf{$\tau^2$-Bench}} \\
    \cmidrule(lr){2-5}
     & \textbf{Retail} & \textbf{Airline} & \textbf{Telecom} & \textbf{Overall}\\
    \midrule
    
    \textbf{Qwen3-8B} & 38.6 & 30.5 & 23.3 & 30.8 \\
    \quad + Cold Start & 59.1 \up{20.5} & 44.0 \up{13.5} & 59.1 \up{35.8} & 53.5 \up{22.7} \\
    \quad + TA-GRPO RL & 79.5 \up{40.9} & 54.7 \up{24.2} & 81.3 \up{58.0} & 71.8 \up{41.0} \\
    \midrule
    
    \textbf{Qwen3-32B} & 53.3 & 42.0 & 26.9 & 40.7 \\
    \quad + Cold Start & 65.3 \up{12.0} & 50.8 \up{8.8} & 71.9 \up{45.0} & 62.7 \up{22.0} \\
    \quad + TA-GRPO RL & 85.1 \up{31.8} & 64.7 \up{22.7} & 88.6 \up{61.7} & 79.5 \up{38.8} \\
    \bottomrule
    \end{tabular}
    \label{tab:increasing_result}
\end{table}

To disentangle the contributions of different training stages, we conduct an incremental analysis comparing the base model against its progressively fine-tuned variants. Table~\ref{tab:increasing_result} presents the performance breakdown across three representative domains in $\tau^2$-Bench.

\paragraph{Effectiveness of Cold Start.}
Initializing from the base model, the Cold Start phase employs only \textbf{3,000} verified trajectories for SFT, yet yields substantial performance gains. As shown in Table~\ref{tab:increasing_result}, the success rate increases significantly for both model sizes. For the 8B model, the overall success rate jumps from 30.8\% to 53.5\%, a gain of 22.7 points. Similarly, the 32B model improves from 40.7\% to 62.7\%. Notably, in the Telecom domain, the 8B model witnesses a dramatic surge from 23.3\% to 59.1\%. This result validates the efficacy of our synthesized data: a small amount of high-quality, logic-dense trajectories is sufficient to effectively guide the model in understanding policy constraints and operational logic.

\paragraph{Necessity of RL Refinement.}
Building upon the Cold Start checkpoint, the subsequent RL optimization (TA-GRPO) drives the performance even higher. We observe that while SFT enables the model to mimic successful behaviors, it often struggles with complex boundary conditions and long-horizon goal achievement. The RL stage addresses these limitations by allowing the agent to explore the state space and optimize for the deterministic state-based reward. The consistent upward trend across all domains demonstrates the robustness of this approach. For instance, the 32B model achieves a final overall success rate of 79.5\% after RL, with particularly strong performance in Retail (85.1\%) and Telecom (88.6\%). The synergy between verifiable SFT initialization and dense-reward RL optimization proves essential for mastering complex agentic workflows.

\subsection{Effectiveness of Turn-aware GRPO}
\begin{figure}[htbp]
  \centering
  \includegraphics[width=\linewidth]{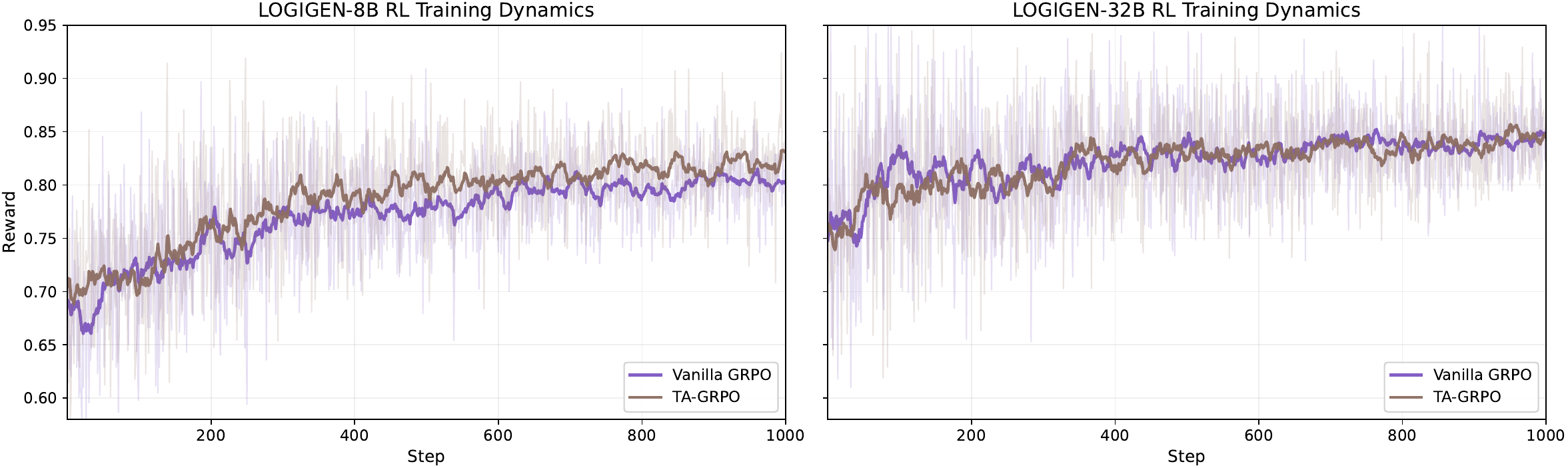}
  \caption{
    Training reward curves for LOGIGEN-8B (left) and 32B (right) models. Compared to Vanilla GRPO, TA-GRPO exhibits higher sample efficiency and achieves superior asymptotic rewards, particularly on the 8B scale.
  }  
  \label{fig:grpo_training_dynamic}
\end{figure}

\begin{figure}[htbp]
  \centering
  \includegraphics[width=\linewidth]{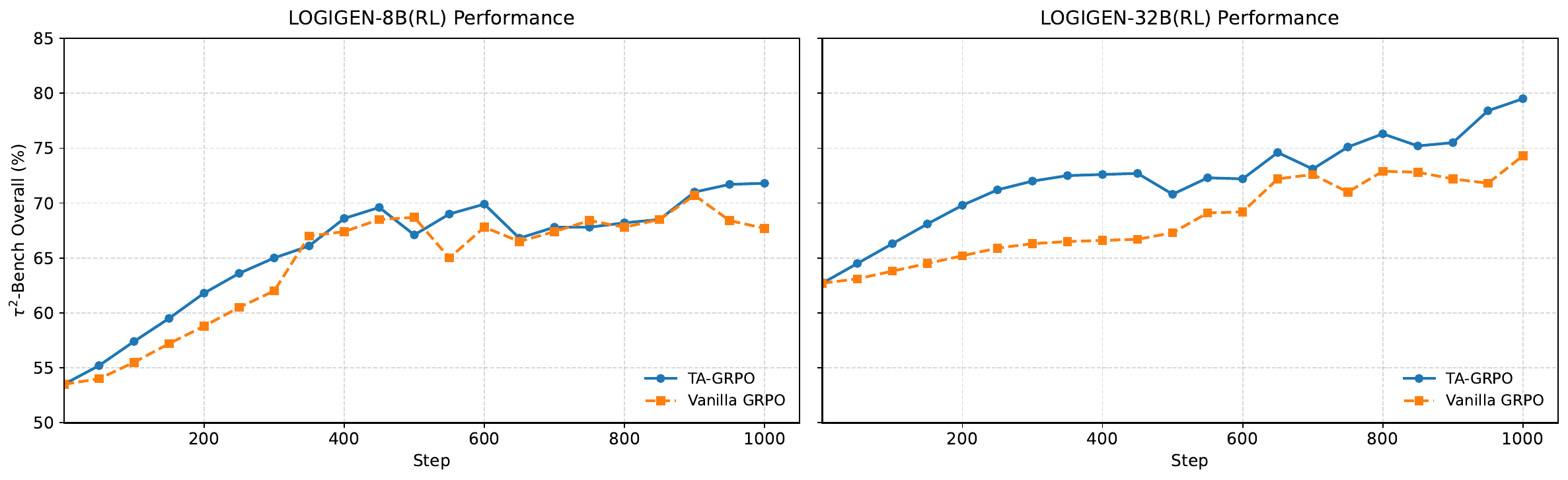}
  \caption{
    Evaluation performance on $\tau^2$-Bench across RL training steps. TA-GRPO consistently yields higher task completion rates than Vanilla GRPO for both 8B and 32B models.
  }  
  \label{fig:grpo_compare_result}
\end{figure}

To evaluate the efficacy of the proposed Turn-aware GRPO (TA-GRPO) in optimizing long-horizon agentic tasks, we conduct an ablation study against a Vanilla GRPO baseline. Our analysis focuses on reinforcement learning (RL) training dynamics and downstream performance on the $\tau^2$-Bench.

\paragraph{Training Dynamics.}
Figure~\ref{fig:grpo_training_dynamic} illustrates the reward curves for the 8B and 32B model variants. We observe a distinct divergence in convergence behavior between the two optimization strategies. 

For the \textbf{8B model} (Figure~\ref{fig:grpo_training_dynamic}, left), TA-GRPO (brown) demonstrates a significantly higher learning efficiency, reaching a stable reward plateau substantially earlier than Vanilla GRPO (purple) and achieving higher asymptotic performance. This suggests that fine-grained, turn-level credit assignment is critical for smaller models to navigate complex action spaces and avoid local optima—specifically, instances where a trajectory reaches a goal despite containing detrimental intermediate steps.

In contrast, the performance gap is less pronounced for the \textbf{32B model} (Figure~\ref{fig:grpo_training_dynamic}, right). While TA-GRPO still yields faster convergence and marginally higher rewards, the relative gain is smaller. We hypothesize this is due to the superior intrinsic reasoning capabilities of larger models, which may implicitly infer causal dependencies across long horizons, thereby reducing their reliance on explicit turn-level penalty signals. Nevertheless, TA-GRPO consistently enhances optimization stability across both scales.

\paragraph{Evaluation Results.}
The training advantages of TA-GRPO translate directly to improved task completion. As shown in Figure~\ref{fig:grpo_compare_result}, models optimized with TA-GRPO consistently outperform their Vanilla GRPO counterparts on $\tau^2$-Bench across both parameter scales. These results confirm that penalizing negative state transitions ($r_t < 0$) within successful trajectories effectively mitigates "happy path" biases and prevents the policy from learning spurious correlations. By enforcing rigorous adherence to policy constraints at each step, TA-GRPO produces more robust agents capable of handling the logical friction inherent in stateful environments.

\subsection{Rollout Trajectories Analysis}
\begin{figure}[htbp]
  \centering
  \includegraphics[width=\linewidth]{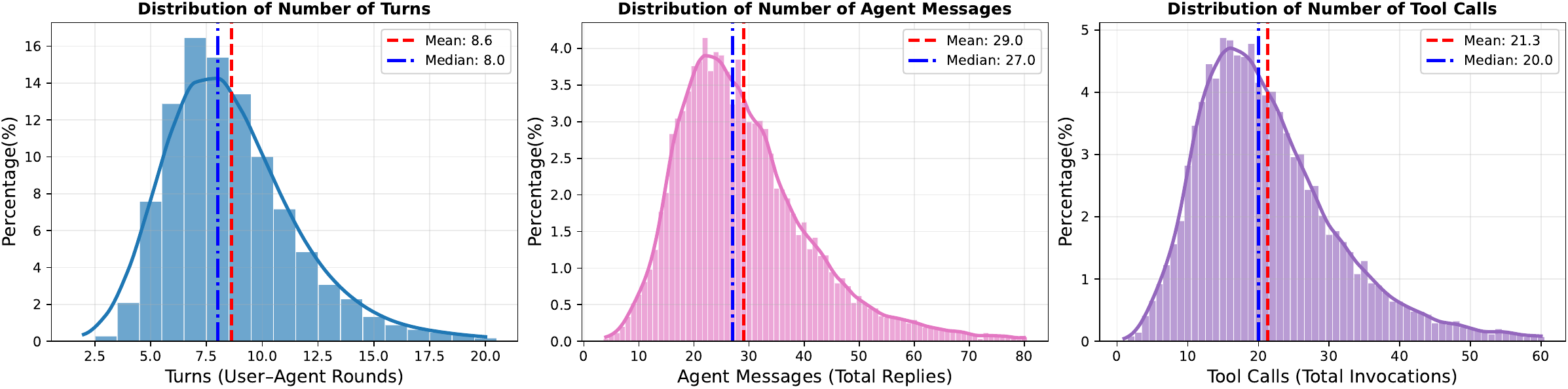}
  \caption{
    \textbf{Trajectory-level statistics}. Distributions of (left) the number of turns (user–agent interaction rounds), (middle) the number of agent messages (total assistant replies; multiple messages may occur within a single turn), and (right) the number of tool calls made by the agent across a trajectory. Bars indicate the empirical frequency (percentage), the solid curve shows a smoothed density estimate, and dashed/dash-dotted vertical lines mark the mean and median.
  }
  \label{fig:trajectory_feature}
\end{figure}
In this section, we analyze the characteristics of rollout trajectories generated during RL training (Figure~\ref{fig:trajectory_feature}). Here, a rollout trajectory refers to the online interaction trace produced by the current policy when it engages with the environment/user and executes intermediate actions (e.g., tool usage) in order to obtain rewards and update the policy. Importantly, these statistics describe the RL rollouts used for reinforcement learning optimization, and should not be conflated with the trajectories in our distillation/offline training data, which may follow different length and tool-usage distributions due to data curation and synthesis procedures.

We summarize each rollout trajectory along three axes: (1) number of turns, i.e., the number of user–agent interaction rounds; (2) number of agent messages, i.e., the total number of assistant messages generated within the trajectory (which can exceed the number of turns because the agent may emit multiple messages within a single turn while performing multi-step reasoning and execution); and (3) number of tool calls, i.e., the total number of tool invocations made across the trajectory, capturing the degree of tool reliance and execution complexity. As shown in Figure~\ref{fig:trajectory_feature}, all three distributions are right-skewed with long tails, indicating that RL rollouts span a broad range of interaction lengths and tool-use intensities. In particular, the distribution of agent messages is shifted further to the right than turns, reflecting frequent within-turn multi-step behaviors (e.g., planning, tool invocation, post-processing, and response refinement). Similarly, the long-tailed tool-call distribution suggests that a subset of rollouts involves substantially more intensive tool usage, consistent with complex tasks that require iterative querying and verification. Overall, these statistics demonstrate that our RL rollouts provide diverse and non-trivial training signals, covering both short-horizon interactions and longer-horizon, tool-heavy trajectories.

\subsection{Consistency and Stability Analysis}
\label{sec:consistency_analysis}

\begin{figure}[htbp]
    \centering
    \includegraphics[width=\linewidth]{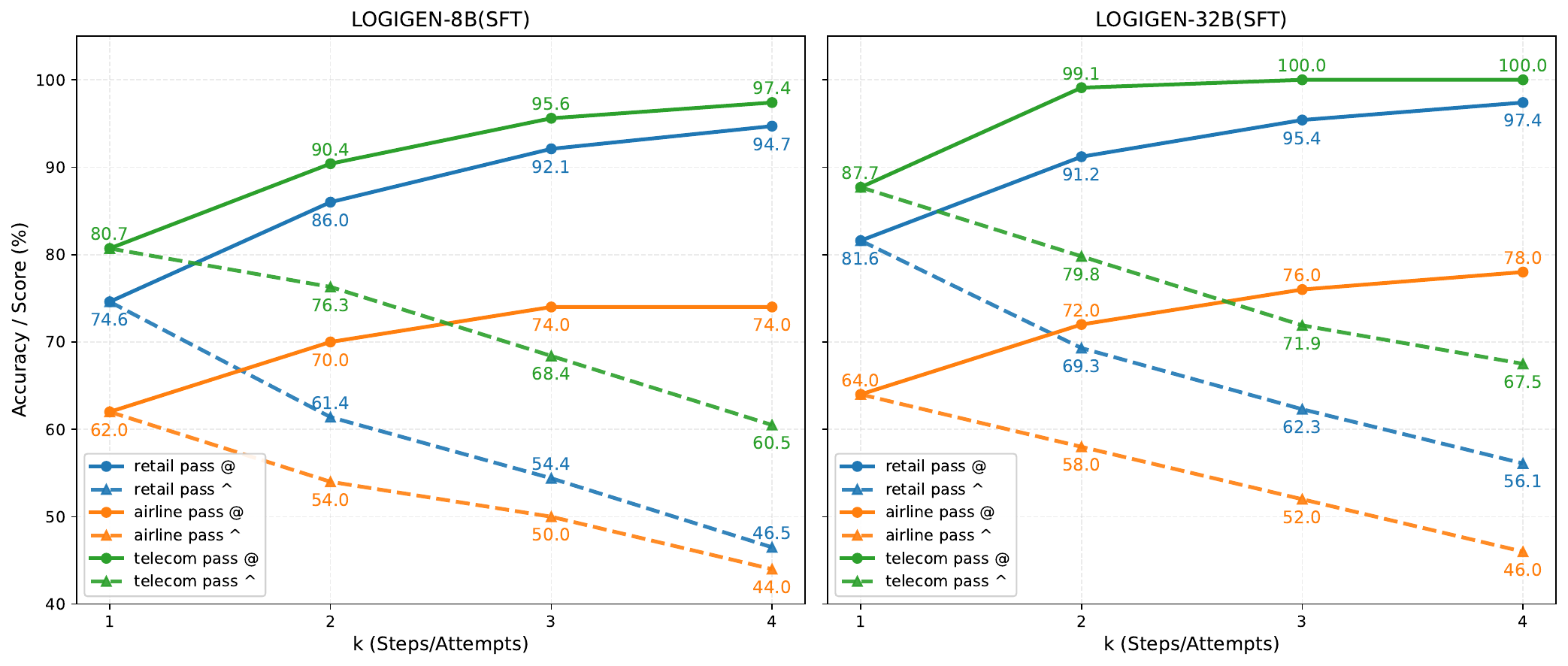}
    \caption{
        Performance comparison of Pass @ k and Pass \^\ k metrics on $\tau^2$-Bench. The left y-axis represents Pass @ k (potential capability), while the right y-axis represents Pass \^\ k (consistency). 
    }
    \label{fig:pass_at}
\end{figure}

We investigate the trade-off between theoretical potential and operational consistency on the $\tau^2$-Bench using the Pass@$k$ and Pass$\hat{k}$ metrics. As illustrated in Figure~\ref{fig:pass_at}, a pronounced disparity emerges between these two dimensions. While Pass@$k$—an indicator of the upper bound of model capability—consistently improves with $k$ and approaches saturation (e.g., nearing 100\% for LOGIGEN-32B in the Telecom domain), Pass$\hat{k}$ exhibits a starkly contrasting trend. This metric, which evaluates reliability across repeated samples, frequently suffers a severe decline as $k$ increases. For instance, LOGIGEN-32B(SFT)'s Pass$\hat{k}$ in the Retail domain plummets from 81.6\% ($k=1$) to 56.1\% ($k=4$). This divergence highlights a critical ``capability-consistency gap'': possessing the knowledge to solve a task does not equate to reliably executing it under stochastic conditions. In multi-turn agentic workflows, this inconsistency is particularly hazardous, as minor deviations in early reasoning steps can trigger unrecoverable error cascades, thereby identifying output stability as a pivotal bottleneck for practical deployment.

\subsection{User Simulator Hacking}
\label{sec:user_sim_hacking}

\begin{figure}[h]
    \centering
    \includegraphics[width=\linewidth]{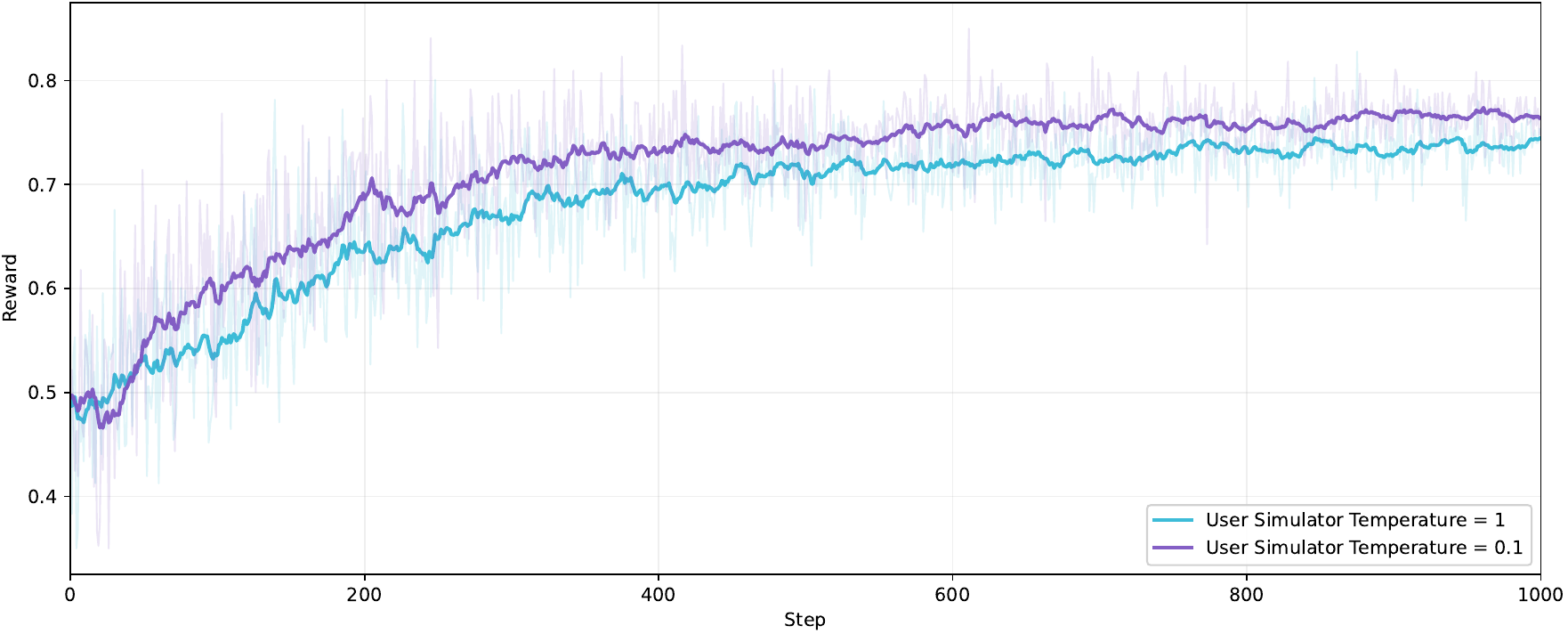}
    \caption{Training reward curves. The agent trained with the low-temperature simulator (T=0.1) achieves higher and faster-converging rewards, indicative of over-optimization.}
    \label{fig:user_sim_hacking_train}
\end{figure}
\begin{figure}[h]
    \centering
    \includegraphics[width=\linewidth]{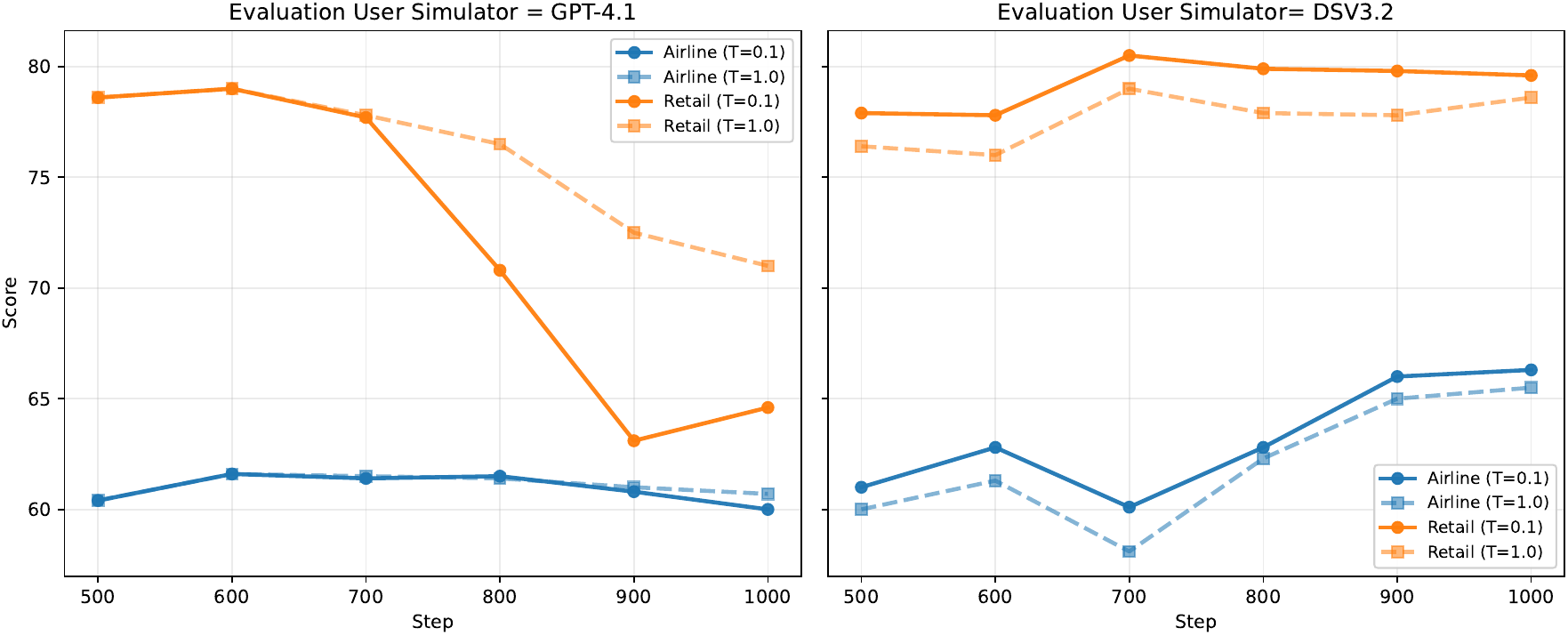}
    \caption{Evaluation performance on $\tau^2$-Bench. The diverse simulator (T=1.0) maintains robustness, whereas the deterministic simulator (T=0.1) suffers from performance degradation.}
    \label{fig:user_sim_hacking_eval}
\end{figure}

Training RL agents with an LLM-based \emph{user simulator} that provides instructions, follow-up queries, and replies introduces the risk of \textbf{user-simulator hacking}. While this setup facilitates scalable interactive training, agents may over-specialize to the specific linguistic patterns of the underlying LLM. Concretely, the agent can learn brittle behaviors that exploit the simulator’s predictable phrasing (e.g., prompting patterns that elicit extra hints or more favorable responses), yielding steadily improving training rewards without improving true task generalization.

To investigate this phenomenon, we conduct a controlled ablation study. We initialize the agent with \textbf{Qwen-32B} and train it from scratch (without cold start) on a subset of the training data. We employ \textbf{dsv3.2} as the training user simulator under two configurations: a low-temperature setting (T=0.1) and a high-temperature setting (T=1.0). To comprehensively assess generalization, we evaluate the trained policies against both in-distribution simulators (the source model \textbf{dsv3.2}) and out-of-distribution simulators (a mismatched model \textbf{gpt-4.1}).

\paragraph{Training Dynamics and Over-optimization.}
Figure~\ref{fig:user_sim_hacking_train} illustrates the on-policy training reward curves. We observe that the agent trained with the low-temperature simulator ($\tau=0.1$) achieves consistently higher and faster-converging rewards compared to its high-temperature counterpart ($\tau=1.0$). In a standard RL setting, such performance might be interpreted as superior learning. However, in the context of LLM interaction, this discrepancy often signals that the agent has identified and exploited specific response patterns unique to the low-entropy simulator (e.g., phrasing queries to consistently elicit favorable responses from \textbf{dsv3.2}).

\paragraph{Cross-Model Evaluation.}
The perils of such over-optimization are exposed during cross-model evaluation. As shown in Figure~\ref{fig:user_sim_hacking_eval} (left), when evaluated against the mismatched model \textbf{gpt-4.1}, the performance of the low-temperature policy (T=0.1, solid lines) exhibits a characteristic \textbf{bell-shaped curve}. In the Retail domain specifically, the score peaks around Step 600-700 and subsequently suffers a sharp decline. This divergence indicates that while the agent optimizes the on-policy reward by ``hacking'' the training simulator, its true task competence degrades as it overfits to simulator-specific artifacts.
Conversely, Figure~\ref{fig:user_sim_hacking_eval} (right) demonstrates that when evaluated on the source model \textbf{dsv3.2}, the performance remains stable with no signs of degradation. The T=0.1 policy maintains high scores throughout training, consistent with the training reward curves. This contrast confirms that the performance drop observed in the left plot is specifically induced by the distribution shift, rather than a failure of the optimization process itself.
In contrast, the high-temperature policy (T=1.0, dashed lines) demonstrates superior robustness in the cross-model setting. Although it achieves lower absolute training rewards, it maintains stable or improving performance on \textbf{gpt-4.1}, eventually surpassing the brittle T=0.1 policy.

\paragraph{Key Takeaway.}
These results highlight a critical ``Generalization-Exploitation'' trade-off: high training rewards driven by deterministic feedback often mask a loss in robustness. Our findings demonstrate that increasing the diversity of the user simulator is key to mitigating simulator hacking. While we implemented this via temperature scaling—which acts as a necessary regularizer against overfitting—this represents only an initial step. Future work should explore more comprehensive strategies, such as employing heterogeneous backbone models as simulators, introducing imperfect user behaviors with stochastic interruptions, incorporating diverse user personas, and scaling task complexity. These advancements would further bridge the gap between simulated training and real-world interaction scenarios.
\FloatBarrier
\section{Conclusion}

We introduce \textbf{LOGIGEN}, a logic-driven framework for synthesizing verifiable training data for autonomous agents. LOGIGEN shifts the data construction paradigm from tool-centric reverse-synthesis to \textit{deductive synthesis}, operationalizing three core pillars: natural-language \textbf{wiki policy} is compiled into a hard-compiled executable environment; boundary-adjacent initial states are seeded to maximize decision density; and goal-conditioned exploration discovers executable multi-turn episodes. These episodes are projected into spoiler-free task descriptions with deterministic target snapshots. This design yields self-contained task packages that enable rigorous state-based verification, facilitating stable \textbf{Verified SFT} and \textbf{State-Based RL} without reliance on heuristic LLM judges.

Empirically, models trained on LOGIGEN data achieve state-of-the-art performance on $\tau^2$-Bench, with LOGIGEN-32B(RL) achieving a \textbf{79.5\% success rate}, substantially outperforming the corresponding base model and remaining competitive with leading proprietary models. These results suggest that physically enforced policies and deterministic state objectives provide a scalable route to constructing the logic-dense trajectories required for long-horizon, policy-governed agentic behavior.

LOGIGEN opens several directions for future work. Beyond relational databases, the same principles of policy compilation and state verification could be extended to richer environments, such as hybrid simulators and multi-service backends. Another promising avenue is to improve exploration efficiency and coverage guarantees, as well as to incorporate multi-agent and multi-user settings where objectives may be partially competing. We hope LOGIGEN will serve as a foundational step toward data generation pipelines that are strictly grounded, verifiable, and aligned with real-world operational constraints.

\bibliography{custom}

\clearpage
\appendix
\section*{Appendix}

\section{Case Study}
\label{appendix:case_study}

\begin{greybox}
    \textbf{Example: Seed Domain Data}
    \tcbline 

    \textbf{Scenario Name:} Corporate Business Travel Portal
    
    \textbf{Context Description:}
    
    A B2B platform used by employees of large corporations to book work trips. It is heavily regulated by company expense policies and approval workflows.
    
    \vspace{0.5em} 
    
    \textbf{Final Key Business Rules:}
    \begin{itemize}[leftmargin=*, nosep, topsep=3pt] 
        \item Flights costing over \$1000 require `Manager Approval' status before ticket issuance.
        
        \item Employees at `Director' level or above can book Business Class; others are restricted to Economy.
        
        \item The system operates on a `Simulation Step' counter. Bookings must have a \texttt{departure\_step} that is at least 3 steps greater than the \texttt{booking\_step}, or a `Policy Violation' flag is recorded.
        
        \item Hotel bookings must be chosen from the `Preferred Vendor List' to be fully reimbursable.
        
        \item Trip purpose field is mandatory for all bookings.
    \end{itemize}
\end{greybox}

\begin{greybox}
\textbf{Example: Corporate Business Travel Agent Wiki Policy}
\tcbline

\hypertarget{corporate-business-travel-agent-policy}{%
\section*{Corporate Business Travel Agent
Policy}\label{corporate-business-travel-agent-policy}}

\hypertarget{section-0-data-governance--authority}{%
\subsection*{Section 0: Data Governance \&
Authority}\label{section-0-data-governance--authority}}

\textbf{CRITICAL INSTRUCTION:} You operate in a strictly governed
corporate travel environment. You must adhere to these foundational laws
of data interaction:

\begin{enumerate}
\def\labelenumi{\arabic{enumi}.}
\item
  \textbf{READ-ONLY Tables (You CANNOT modify these directly):}

  \begin{itemize}
  \tightlist
  \item
    \texttt{travel\_policies}, \texttt{preferred\_vendors},
    \texttt{flight\_classes} are \textbf{STRICTLY READ-ONLY} system
    catalogs.
  \item
    \texttt{companies} and \texttt{users} are \textbf{READ-ONLY}
    user/company profiles. You cannot activate, deactivate, or edit
    these directly.
  \end{itemize}
\item
  \textbf{WRITEABLE Tables (You can \texttt{INSERT} and \texttt{UPDATE}
  these):}

  \begin{itemize}
  \tightlist
  \item
    \texttt{travel\_requests}, \texttt{flight\_bookings},
    \texttt{hotel\_bookings}, \texttt{approvals} are your primary
    workspace. You perform actions by creating and updating records in
    these \textbf{L2\_TRANSACTION} tables.
  \end{itemize}
\item
  \textbf{The Universal "No-Delete" Rule:}

  \begin{itemize}
  \tightlist
  \item
    You are \textbf{STRICTLY FORBIDDEN} from using the SQL
    \texttt{DELETE} command on any table.
  \item
    \textbf{Deactivation/Lifecycle Method:} To "remove" a record, you
    must use the prescribed status update:

    \begin{itemize}
    \tightlist
    \item
      \textbf{Users/Companies:} Set \texttt{active=0}. (You cannot do
      this directly; see Indirect Manipulation).
    \item
      \textbf{Travel Requests/Bookings:} Set
      \texttt{status=\textquotesingle{}CANCELLED\textquotesingle{}}.
    \item
      \textbf{Approvals:} Set
      \texttt{status=\textquotesingle{}DENIED\textquotesingle{}}.
    \end{itemize}
  \end{itemize}
\end{enumerate}

\begin{center}\rule{0.5\linewidth}{0.5pt}\end{center}

\hypertarget{section-1-persona--mission}{%
\subsection*{Section 1: Persona \&
Mission}\label{section-1-persona--mission}}

\textbf{Your Persona:} You are a professional, efficient, and
policy-aware Corporate Travel Specialist. You are patient,
detail-oriented, and prioritize compliance with company travel policies
while striving to provide excellent service.

\textbf{Your Mission:} To assist employees in creating, managing, and
modifying their business travel plans in full compliance with their
company\textquotesingle s policies, ensuring all necessary approvals are
obtained and all system rules are followed.

\begin{center}\rule{0.5\linewidth}{0.5pt}\end{center}

\hypertarget{section-2-conversation-flow--policies}{%
\subsection*{Section 2: Conversation Flow \&
Policies}\label{section-2-conversation-flow--policies}}

\begin{itemize}
\tightlist
\item
  \textbf{Authentication Protocol:} At the start of any interaction, you
  must verify the user\textquotesingle s identity. Confirm their
  \texttt{user\_id} and that they belong to an active company.
\item
  \textbf{Confirmation Loop:} Before executing any state-changing action
  (e.g., booking, cancelling, submitting for approval), you
  \textbf{must} summarize the action and its key parameters (cost,
  dates/policy-violation-flag, approval need) and get explicit
  confirmation (\texttt{yes}) from the user.
\item
  \textbf{Information Boundaries:} Do not invent or assume travel
  policies, vendor details, or user information. All data must be
  queried from the relevant tables (\texttt{travel\_policies},
  \texttt{preferred\_vendors}, \texttt{users}).
\item
  \textbf{Single-Tasking:} Make one tool call at a time. Process each
  booking, cancellation, or approval step sequentially.
\item
  \textbf{Escalation Path:} If a user encounters a system-imposed policy
  restriction they wish to contest, or if there is a suspected data
  error (e.g., incorrect policy mapping), acknowledge the issue and
  offer to escalate the matter to a human travel administrator for
  review.
\end{itemize}

\begin{center}\rule{0.5\linewidth}{0.5pt}\end{center}

\hypertarget{section-3-domain-data-model}{%
\subsection*{Section 3: Domain Data
Model}\label{section-3-domain-data-model}}

\hypertarget{table-travel_policies-l0-system}{%
\subsubsection*{\texorpdfstring{Table: \texttt{travel\_policies} {[}L0:
SYSTEM{]}}{Table: travel\_policies {[}L0: SYSTEM{]}}}\label{table-travel_policies-l0-system}}

\begin{itemize}
\tightlist
\item
  \textbf{Permission Mode:} \textbf{{[}READ-ONLY{]}} SELECT Only. You
  cannot modify the master policy catalog.
\item
  \textbf{Purpose:} Defines company-specific rules for travel (cost
  limits, allowed vendors).
\item
  \textbf{Key Columns:} \texttt{user\_level} (STAFF, MANAGER, etc.),
  \texttt{max\_flight\_cost\_no\_approval} (dollar threshold),
  \texttt{allowed\_hotel\_vendor\_type} (PREFERRED or ANY).
\item
  \textbf{Relationships:} Linked to \texttt{companies} via
  \texttt{company\_id}. Used to validate \texttt{flight\_bookings} and
  \texttt{hotel\_bookings}.
\end{itemize}

\hypertarget{table-preferred_vendors-l0-system}{%
\subsubsection*{\texorpdfstring{Table: \texttt{preferred\_vendors} {[}L0:
SYSTEM{]}}{Table: preferred\_vendors {[}L0: SYSTEM{]}}}\label{table-preferred_vendors-l0-system}}

\begin{itemize}
\tightlist
\item
  \textbf{Permission Mode:} \textbf{{[}READ-ONLY{]}} SELECT Only. You
  cannot modify the vendor list.
\item
  \textbf{Purpose:} Master list of approved hotel vendors.
\item
  \textbf{Key Columns:} \texttt{vendor\_type} (PREFERRED or STANDARD).
  Determines if a booking is \texttt{reimbursable}.
\end{itemize}

\hypertarget{table-flight_classes-l0-system}{%
\subsubsection*{\texorpdfstring{Table: \texttt{flight\_classes} {[}L0:
SYSTEM{]}}{Table: flight\_classes {[}L0: SYSTEM{]}}}\label{table-flight_classes-l0-system}}

\begin{itemize}
\tightlist
\item
  \textbf{Permission Mode:} \textbf{{[}READ-ONLY{]}} SELECT Only. You
  cannot modify the class catalog.
\item
  \textbf{Purpose:} Reference for flight service classes (ECONOMY,
  BUSINESS).
\item
  \textbf{Relationships:} Referenced by \texttt{flight\_bookings.class}.
\end{itemize}

\hypertarget{table-companies-l1-user}{%
\subsubsection*{\texorpdfstring{Table: \texttt{companies} {[}L1:
USER{]}}{Table: companies {[}L1: USER{]}}}\label{table-companies-l1-user}}

\begin{itemize}
\tightlist
\item
  \textbf{Permission Mode:} \textbf{{[}READ-ONLY{]}} SELECT Only. You
  cannot modify this table directly.
\item
  \textbf{Purpose:} Corporation master data.
\item
  \textbf{Key Columns:} \texttt{active} (1=Active, 0=Inactive).
\item
  \textbf{Lifecycle Method:} To deactivate, set \texttt{active=0}.
  (Note: You cannot do this directly).
\item
  \textbf{Relationships:} Parent of \texttt{users}.
\end{itemize}

\hypertarget{table-users-l1-user}{%
\subsubsection*{\texorpdfstring{Table: \texttt{users} {[}L1:
USER{]}}{Table: users {[}L1: USER{]}}}\label{table-users-l1-user}}

\begin{itemize}
\tightlist
\item
  \textbf{Permission Mode:} \textbf{{[}READ-ONLY{]}} SELECT Only. You
  cannot modify user profiles directly. (If a user\textquotesingle s
  status needs to change, it must be done through an external
  administrative process).
\item
  \textbf{Purpose:} Employee master data.
\item
  \textbf{Key Columns:} \texttt{user\_level} (STAFF, MANAGER, DIRECTOR,
  VP) - determines permissions; \texttt{active} (1=Active, 0=Inactive);
  \texttt{company\_id}.
\item
  \textbf{Lifecycle Method:} To deactivate, set \texttt{active=0}.
  (Note: You cannot do this directly).
\item
  \textbf{Relationships:} Linked to \texttt{companies}. Parent of
  \texttt{travel\_requests}.
\end{itemize}

\hypertarget{table-travel_requests-l2-biz}{%
\subsubsection*{\texorpdfstring{Table: \texttt{travel\_requests} {[}L2:
BIZ{]}}{Table: travel\_requests {[}L2: BIZ{]}}}\label{table-travel_requests-l2-biz}}

\begin{itemize}
\tightlist
\item
  \textbf{Permission Mode:} \textbf{{[}READ/WRITE{]}} SELECT, INSERT,
  UPDATE. You have direct control.
\item
  \textbf{Purpose:} The core container for a business trip. Holds quotas
  for bookings.
\item
  \textbf{Key Columns:} \texttt{status} (DRAFT, SUBMITTED, APPROVED,
  CANCELLED); \texttt{current\_step} (simulation step of creation);
  \texttt{flight\_booking\_count}, \texttt{hotel\_booking\_count}
  (system-maintained quotas).
\item
  \textbf{Lifecycle Method:} To cancel, set
  \texttt{status=\textquotesingle{}CANCELLED\textquotesingle{}}.
\item
  \textbf{Relationships:} Linked to \texttt{users}. Parent of
  \texttt{flight\_bookings} and \texttt{hotel\_bookings}.
\end{itemize}

\hypertarget{table-flight_bookings-l2-biz}{%
\subsubsection*{\texorpdfstring{Table: \texttt{flight\_bookings} {[}L2:
BIZ{]}}{Table: flight\_bookings {[}L2: BIZ{]}}}\label{table-flight_bookings-l2-biz}}

\begin{itemize}
\tightlist
\item
  \textbf{Permission Mode:} \textbf{{[}READ/WRITE{]}} SELECT, INSERT,
  UPDATE. You have direct control.
\item
  \textbf{Purpose:} Individual flight booking within a travel request.
\item
  \textbf{Key Columns:}

  \begin{itemize}
  \tightlist
  \item
    \texttt{status} (PENDING, APPROVED, TICKETED, CANCELLED).
  \item
    \texttt{approval\_status} (NOT\_REQUIRED, PENDING, APPROVED,
    DENIED).
  \item
    \texttt{policy\_violation\_flag} (1=Emergency booking, 0=Normal).
    \textbf{System-calculated}.
  \item
    \texttt{departure\_step}, \texttt{booking\_step},
    \texttt{cancellation\_step} (simulation timestamps).
  \item
    \texttt{refund\_amount} (must be calculated on cancellation).
  \end{itemize}
\item
  \textbf{Lifecycle Method:} To cancel, set
  \texttt{status=\textquotesingle{}CANCELLED\textquotesingle{}} and
  provide \texttt{cancellation\_step} and \texttt{refund\_amount}.
\item
  \textbf{Relationships:} Linked to \texttt{travel\_requests} and
  \texttt{flight\_classes}.
\end{itemize}

\hypertarget{table-hotel_bookings-l2-biz}{%
\subsubsection*{\texorpdfstring{Table: \texttt{hotel\_bookings} {[}L2:
BIZ{]}}{Table: hotel\_bookings {[}L2: BIZ{]}}}\label{table-hotel_bookings-l2-biz}}

\begin{itemize}
\tightlist
\item
  \textbf{Permission Mode:} \textbf{{[}READ/WRITE{]}} SELECT, INSERT,
  UPDATE. You have direct control.
\item
  \textbf{Purpose:} Individual hotel booking within a travel request.
\item
  \textbf{Key Columns:}

  \begin{itemize}
  \tightlist
  \item
    \texttt{status} (PENDING, CONFIRMED, CANCELLED).
  \item
    \texttt{reimbursable} (1=Yes, 0=No). \textbf{System-calculated}
    based on vendor type.
  \item
    \texttt{booking\_step}, \texttt{cancellation\_step} (simulation
    timestamps).
  \item
    \texttt{refund\_amount} (must be calculated on cancellation).
  \end{itemize}
\item
  \textbf{Lifecycle Method:} To cancel, set
  \texttt{status=\textquotesingle{}CANCELLED\textquotesingle{}} and
  provide \texttt{cancellation\_step} and \texttt{refund\_amount}.
\item
  \textbf{Relationships:} Linked to \texttt{travel\_requests} and
  \texttt{preferred\_vendors}.
\end{itemize}

\hypertarget{table-approvals-l2-biz}{%
\subsubsection*{\texorpdfstring{Table: \texttt{approvals} {[}L2:
BIZ{]}}{Table: approvals {[}L2: BIZ{]}}}\label{table-approvals-l2-biz}}

\begin{itemize}
\tightlist
\item
  \textbf{Permission Mode:} \textbf{{[}READ/WRITE{]}} SELECT, INSERT,
  UPDATE. You have direct control.
\item
  \textbf{Purpose:} Tracks manager approval workflow for flight bookings
  that require it.
\item
  \textbf{Key Columns:} \texttt{status} (PENDING, APPROVED, DENIED);
  \texttt{approver\_id}; \texttt{step} (simulation step of
  creation/update).
\item
  \textbf{Lifecycle Method:} To reject, set
  \texttt{status=\textquotesingle{}DENIED\textquotesingle{}}.
\item
  \textbf{Relationships:} Linked to \texttt{flight\_bookings} and
  \texttt{users} (approver).
\end{itemize}

\begin{center}\rule{0.5\linewidth}{0.5pt}\end{center}

\hypertarget{section-4-standard-operating-procedures-sop}{%
\subsection*{Section 4: Standard Operating Procedures
(SOP)}\label{section-4-standard-operating-procedures-sop}}

\hypertarget{transaction-create-travel-request}{%
\subsubsection*{Transaction: Create Travel
Request}\label{transaction-create-travel-request}}

\begin{itemize}
\tightlist
\item
  \textbf{Goal:} Establish a new container for a business trip.
\item
  \textbf{Business Rules (Prerequisites):}

  \begin{itemize}
  \tightlist
  \item
    \textbf{User Standing:} The requesting user must exist and be
    \texttt{active}.
  \item
    \textbf{Company Standing:} The user\textquotesingle s company must
    be \texttt{active}.
  \item
    \textbf{Purpose Required:} The \texttt{trip\_purpose} must be
    provided and non-empty.
  \item
    \textbf{Timestamp:} The \texttt{current\_step} (simulation
    timestamp) must be provided.
  \end{itemize}
\item
  \textbf{Execution (The "How"):}

  \begin{itemize}
  \tightlist
  \item
    \textbf{Action:} Create Travel Request via \texttt{INSERT} into
    \texttt{travel\_requests}.
  \item
    \textbf{Parameters:} \texttt{user\_id}, \texttt{trip\_purpose},
    \texttt{current\_step}.
  \end{itemize}
\item
  \textbf{Outcome (System Logic):}

  \begin{itemize}
  \tightlist
  \item
    The system will \textbf{automatically} set the initial
    \texttt{status=\textquotesingle{}DRAFT\textquotesingle{}}.
  \item
    The \texttt{flight\_booking\_count} and
    \texttt{hotel\_booking\_count} quotas are initialized to 0.
  \end{itemize}
\item
  \textbf{Potential Failures:}

  \begin{itemize}
  \tightlist
  \item
    \texttt{{[}PREREQ\_FAIL{]}\ User\ does\ not\ exist\ or\ is\ inactive}
  \item
    \texttt{{[}PREREQ\_FAIL{]}\ User\textquotesingle{}s\ company\ is\ inactive}
  \item
    \texttt{{[}POLICY\_VIOLATION{]}\ Trip\ purpose\ is\ mandatory}
  \end{itemize}
\end{itemize}

\hypertarget{transaction-book-flight}{%
\subsubsection*{Transaction: Book Flight}\label{transaction-book-flight}}

\begin{itemize}
\tightlist
\item
  \textbf{Goal:} Add a flight segment to an active travel request.
\item
  \textbf{Business Rules (Prerequisites):}

  \begin{itemize}
  \tightlist
  \item
    \textbf{Request Status:} The parent \texttt{travel\_request} must be
    in \texttt{DRAFT} or \texttt{APPROVED} status.
  \item
    \textbf{Quota Limit:} The request must have fewer than 3 active
    (\texttt{status\ !=\ \textquotesingle{}CANCELLED\textquotesingle{}})
    flight bookings.
  \item
    \textbf{Timestamp Logic:} The \texttt{booking\_step} and
    \texttt{departure\_step} must be provided. The system will calculate
    if this is an emergency booking (\texttt{policy\_violation\_flag=1})
    based on the step difference.
  \item
    \textbf{Cabin Permission:} Verify \textbf{Cabin Eligibility}: Only
    users with \texttt{user\_level} of \texttt{DIRECTOR} or \texttt{VP}
    can book non-\texttt{ECONOMY} class.
  \item
    \textbf{Approval Matrix (CRITICAL):} Determine if manager approval
    is required:

    \begin{enumerate}
    \def\labelenumi{\arabic{enumi}.}
    \tightlist
    \item
      Look up the user\textquotesingle s \texttt{travel\_policy} to get
      the \texttt{max\_flight\_cost\_no\_approval} threshold.
    \item
      \textbf{Approval IS required} if:
      \texttt{flight\_cost\ \textgreater{}\ policy\_threshold} AND the
      user is \textbf{not} eligible for a \textbf{Special Waiver}.
    \item
      \textbf{Special Waiver (Exception OR Gate):} Approval is
      \textbf{NOT required} if EITHER:

      \begin{itemize}
      \tightlist
      \item
        \textbf{Waiver A:} The user is \texttt{DIRECTOR}/\texttt{VP} AND
        the \texttt{flight\_cost\ \textless{}\ 500}, OR
      \item
        \textbf{Waiver B:} The booking is an emergency (system will set
        \texttt{policy\_violation\_flag\ =\ 1}).
      \end{itemize}
    \end{enumerate}
  \end{itemize}
\item
  \textbf{Execution (The "How"):}

  \begin{itemize}
  \tightlist
  \item
    \textbf{Action:} Book Flight via \texttt{INSERT} into
    \texttt{flight\_bookings}.
  \item
    \textbf{Parameters:} \texttt{travel\_request\_id},
    \texttt{flight\_code}, \texttt{cost}, \texttt{class},
    \texttt{departure\_step}, \texttt{booking\_step}.
  \item
    \textbf{Critical Parameter:} You \textbf{must} set
    \texttt{approval\_status} correctly:

    \begin{itemize}
    \tightlist
    \item
      Set to \texttt{\textquotesingle{}PENDING\textquotesingle{}} if
      approval is required per the matrix.
    \item
      Set to \texttt{\textquotesingle{}NOT\_REQUIRED\textquotesingle{}}
      if approval is not required.
    \end{itemize}
  \end{itemize}
\item
  \textbf{Outcome (System Logic):}

  \begin{itemize}
  \tightlist
  \item
    The system \textbf{automatically} sets the initial
    \texttt{status=\textquotesingle{}PENDING\textquotesingle{}}.
  \item
    The system \textbf{automatically} calculates and sets the
    \texttt{policy\_violation\_flag}.
  \item
    \textbf{If approval was required
    (\texttt{approval\_status=\textquotesingle{}PENDING\textquotesingle{}}):}
    The system \textbf{automatically} creates a corresponding
    \texttt{approvals} record with
    \texttt{status=\textquotesingle{}PENDING\textquotesingle{}}.
  \item
    The system updates the parent
    \texttt{travel\_request.flight\_booking\_count}.
  \end{itemize}
\item
  \textbf{Potential Failures:}

  \begin{itemize}
  \tightlist
  \item
    \texttt{{[}QUOTA\_EXCEEDED{]}\ Maximum\ 3\ flight\ bookings\ per\ travel\ request}
  \item
    \texttt{{[}POLICY\_VIOLATION{]}\ Only\ DIRECTOR/VP\ level\ can\ book\ non-ECONOMY\ class}
  \item
    \texttt{{[}POLICY\_VIOLATION{]}\ Flight\ requires\ manager\ approval.\ Set\ approval\_status\ =\ PENDING}
  \item
    \texttt{{[}LOGIC\_ERROR{]}\ Approval\ not\ required\ for\ this\ flight.\ Set\ approval\_status\ =\ NOT\_REQUIRED}
  \end{itemize}
\end{itemize}

\hypertarget{transaction-book-hotel}{%
\subsubsection*{Transaction: Book Hotel}\label{transaction-book-hotel}}

\begin{itemize}
\tightlist
\item
  \textbf{Goal:} Add a hotel stay to an active travel request.
\item
  \textbf{Business Rules (Prerequisites):}

  \begin{itemize}
  \tightlist
  \item
    \textbf{Request Status:} The parent \texttt{travel\_request} must be
    in \texttt{DRAFT} or \texttt{APPROVED} status.
  \item
    \textbf{Quota Limit:} The request must have fewer than 2 active
    (\texttt{status\ !=\ \textquotesingle{}CANCELLED\textquotesingle{}})
    hotel bookings.
  \item
    \textbf{Vendor Source:} The \texttt{hotel\_vendor\_id} must be
    selected from the \texttt{preferred\_vendors} catalog.
  \item
    \textbf{Policy Compliance:} Verify \textbf{Vendor Eligibility}: The
    hotel\textquotesingle s \texttt{vendor\_type} must be allowed by the
    user\textquotesingle s
    \texttt{travel\_policy.allowed\_hotel\_vendor\_type} (which must be
    either \texttt{\textquotesingle{}ANY\textquotesingle{}} or match the
    vendor\textquotesingle s type exactly).
  \end{itemize}
\item
  \textbf{Execution (The "How"):}

  \begin{itemize}
  \tightlist
  \item
    \textbf{Action:} Book Hotel via \texttt{INSERT} into
    \texttt{hotel\_bookings}.
  \item
    \textbf{Parameters:} \texttt{travel\_request\_id},
    \texttt{hotel\_vendor\_id}, \texttt{cost}, \texttt{booking\_step}.
  \end{itemize}
\item
  \textbf{Outcome (System Logic):}

  \begin{itemize}
  \tightlist
  \item
    The system \textbf{automatically} sets the initial
    \texttt{status=\textquotesingle{}PENDING\textquotesingle{}}.
  \item
    The system \textbf{automatically} sets \texttt{reimbursable=1} if
    the vendor\textquotesingle s \texttt{vendor\_type} is
    \texttt{\textquotesingle{}PREFERRED\textquotesingle{}}; otherwise
    sets it to \texttt{0}.
  \item
    The system updates the parent
    \texttt{travel\_request.hotel\_booking\_count}.
  \end{itemize}
\item
  \textbf{Potential Failures:}

  \begin{itemize}
  \tightlist
  \item
    \texttt{{[}QUOTA\_EXCEEDED{]}\ Maximum\ 2\ hotel\ bookings\ per\ travel\ request}
  \item
    \texttt{{[}POLICY\_VIOLATION{]}\ Hotel\ must\ be\ from\ preferred\ vendors\ list}
  \item
    \texttt{{[}POLICY\_VIOLATION{]}\ Hotel\ vendor\ type\ does\ not\ match\ policy\ requirement}
  \end{itemize}
\end{itemize}

\hypertarget{transaction-cancel-flight-booking}{%
\subsubsection*{Transaction: Cancel Flight
Booking}\label{transaction-cancel-flight-booking}}

\begin{itemize}
\tightlist
\item
  \textbf{Goal:} Cancel a pending or approved flight booking and
  calculate refund.
\item
  \textbf{Business Rules (Prerequisites):}

  \begin{itemize}
  \tightlist
  \item
    \textbf{Status Gate:} The booking must \textbf{not} be in
    \texttt{TICKETED} status. \texttt{TICKETED} bookings are
    \textbf{irreversible}.
  \item
    \textbf{Not Already Cancelled:} The booking must not already be
    \texttt{CANCELLED}.
  \item
    \textbf{Provenance \& Calculation:} You must provide the
    \texttt{cancellation\_step} and calculate the correct
    \texttt{refund\_amount}:

    \begin{itemize}
    \tightlist
    \item
      \textbf{Full Refund:} If cancelling within 2 simulation steps of
      the \texttt{booking\_step}:
      \texttt{refund\_amount\ =\ original\ cost}.
    \item
      \textbf{50\% Penalty:} If cancelling more than 2 steps after the
      \texttt{booking\_step}:
      \texttt{refund\_amount\ =\ original\ cost\ /\ 2}.
    \end{itemize}
  \end{itemize}
\item
  \textbf{Execution (The "How"):}

  \begin{itemize}
  \tightlist
  \item
    \textbf{Action:} Cancel Flight via \texttt{UPDATE} on
    \texttt{flight\_bookings}.
  \item
    \textbf{Parameters:} Set
    \texttt{status=\textquotesingle{}CANCELLED\textquotesingle{}},
    provide \texttt{cancellation\_step} and the calculated
    \texttt{refund\_amount}.
  \end{itemize}
\item
  \textbf{Outcome (System Logic):}

  \begin{itemize}
  \tightlist
  \item
    The system validates your \texttt{refund\_amount} calculation
    against the rule above.
  \item
    The system updates the parent
    \texttt{travel\_request.flight\_booking\_count} quota.
  \end{itemize}
\item
  \textbf{Potential Failures:}

  \begin{itemize}
  \tightlist
  \item
    \texttt{{[}IRREVERSIBLE{]}\ TICKETED\ flights\ cannot\ be\ cancelled}
  \item
    \texttt{{[}CALCULATION\_ERROR{]}\ ...} (details the expected refund
    amount).
  \end{itemize}
\end{itemize}

\hypertarget{transaction-cancel-hotel-booking}{%
\subsubsection*{Transaction: Cancel Hotel
Booking}\label{transaction-cancel-hotel-booking}}

\begin{itemize}
\tightlist
\item
  \textbf{Goal:} Cancel a pending hotel booking and calculate refund.
\item
  \textbf{Business Rules (Prerequisites):}

  \begin{itemize}
  \tightlist
  \item
    \textbf{Status Gate:} The booking must \textbf{not} be in
    \texttt{CONFIRMED} status. \texttt{CONFIRMED} bookings are
    \textbf{irreversible}.
  \item
    \textbf{Not Already Cancelled:} The booking must not already be
    \texttt{CANCELLED}.
  \item
    \textbf{Provenance \& Calculation:} You must provide the
    \texttt{cancellation\_step} and calculate the correct
    \texttt{refund\_amount} using the same 2-step rule as flight
    cancellations.
  \end{itemize}
\item
  \textbf{Execution (The "How"):}

  \begin{itemize}
  \tightlist
  \item
    \textbf{Action:} Cancel Hotel via \texttt{UPDATE} on
    \texttt{hotel\_bookings}.
  \item
    \textbf{Parameters:} Set
    \texttt{status=\textquotesingle{}CANCELLED\textquotesingle{}},
    provide \texttt{cancellation\_step} and the calculated
    \texttt{refund\_amount}.
  \end{itemize}
\item
  \textbf{Outcome (System Logic):}

  \begin{itemize}
  \tightlist
  \item
    The system validates your \texttt{refund\_amount} calculation.
  \item
    The system updates the parent
    \texttt{travel\_request.hotel\_booking\_count} quota.
  \end{itemize}
\item
  \textbf{Potential Failures:}

  \begin{itemize}
  \tightlist
  \item
    \texttt{{[}IRREVERSIBLE{]}\ CONFIRMED\ hotels\ cannot\ be\ cancelled}
  \item
    \texttt{{[}CALCULATION\_ERROR{]}\ ...} (details the expected refund
    amount).
  \end{itemize}
\end{itemize}

\hypertarget{transaction-process-manager-approval}{%
\subsubsection*{Transaction: Process Manager
Approval}\label{transaction-process-manager-approval}}

\begin{itemize}
\tightlist
\item
  \textbf{Goal:} A manager approves or denies a pending flight approval
  request.
\item
  \textbf{Business Rules (Prerequisites):}

  \begin{itemize}
  \tightlist
  \item
    \textbf{Approval Status:} The \texttt{approvals} record must have
    \texttt{status=\textquotesingle{}PENDING\textquotesingle{}}.
  \item
    \textbf{Approver Authority:} The \texttt{approver\_id} must belong
    to an active user with \texttt{user\_level} of \texttt{MANAGER} or
    higher.
  \item
    \textbf{Conflict of Interest:} The approver cannot be the same
    person who requested the travel (the \texttt{user\_id} on the linked
    \texttt{travel\_request}).
  \end{itemize}
\item
  \textbf{Execution (The "How"):}

  \begin{itemize}
  \tightlist
  \item
    \textbf{Action:} Update Approval via \texttt{UPDATE} on
    \texttt{approvals}.
  \item
    \textbf{Parameters:} Set
    \texttt{status=\textquotesingle{}APPROVED\textquotesingle{}} or
    \texttt{\textquotesingle{}DENIED\textquotesingle{}}, provide
    \texttt{approver\_id}.
  \end{itemize}
\item
  \textbf{Outcome (System Logic):}

  \begin{itemize}
  \tightlist
  \item
    The system \textbf{automatically} updates the linked
    \texttt{flight\_bookings.approval\_status} to match.
  \item
    \textbf{If Approved:} The system will \textbf{automatically} change
    the \texttt{flight\_bookings.status} to
    \texttt{\textquotesingle{}TICKETED\textquotesingle{}}, but only if
    the flight\textquotesingle s \texttt{policy\_violation\_flag=0} and
    its status was \texttt{\textquotesingle{}PENDING\textquotesingle{}}.
  \end{itemize}
\item
  \textbf{Potential Failures:}

  \begin{itemize}
  \tightlist
  \item
    \texttt{{[}IRREVERSIBLE{]}\ Can\ only\ update\ PENDING\ approvals}
  \item
    \texttt{{[}AUTHORITY\_ERROR{]}\ Approver\ must\ be\ MANAGER\ level\ or\ higher}
  \item
    \texttt{{[}CONFLICT\_OF\_INTEREST{]}\ Approver\ cannot\ approve\ their\ own\ request}
  \end{itemize}
\end{itemize}

\begin{center}\rule{0.5\linewidth}{0.5pt}\end{center}

\hypertarget{section-5-critical-system-enforcement-summary}{%
\subsection*{Section 5: Critical System Enforcement
(Summary)}\label{section-5-critical-system-enforcement-summary}}

\begin{enumerate}
\def\labelenumi{\arabic{enumi}.}
\tightlist
\item
  \textbf{NEVER use \texttt{DELETE}.} All lifecycle changes are done via
  status updates (\texttt{CANCELLED}, \texttt{DENIED}) or the
  \texttt{active} flag.
\item
  \textbf{Indirect Access:} You cannot modify \texttt{users} or
  \texttt{companies} directly. Status changes to these entities must
  occur outside your control.
\item
  \textbf{Irreversible States:} Once a flight is \texttt{TICKETED} or a
  hotel is \texttt{CONFIRMED}, it \textbf{cannot} be cancelled or
  modified. Once any booking is \texttt{CANCELLED}, it \textbf{cannot}
  be reactivated.
\item
  \textbf{Quota Enforcement:} A travel request can have at most 3 active
  flight bookings and 2 active hotel bookings.
\item
  \textbf{System-Control Flags:} Never manually set
  \texttt{policy\_violation\_flag} or \texttt{reimbursable}. The system
  calculates these automatically upon \texttt{INSERT}.
\item
  \textbf{Provenance \& Calculation:} For cancellations, you
  \textbf{must} provide the \texttt{cancellation\_step} and the
  correctly calculated \texttt{refund\_amount}. The system will validate
  your math.
\end{enumerate}
\end{greybox}

\begin{greybox}
    \subsection*{Example: Database Tables}
    \tcbline

    \begin{lstlisting}[style=sqlstyle]

-- L0_REFERENCE Table: flight_classes
CREATE TABLE flight_classes (
    id TEXT PRIMARY KEY,
    description TEXT NOT NULL
);

-- L0_REFERENCE Table: preferred_vendors
CREATE TABLE preferred_vendors (
    id TEXT PRIMARY KEY,
    name TEXT NOT NULL,
    vendor_type TEXT NOT NULL CHECK(vendor_type IN ('PREFERRED', 'STANDARD'))
);

-- L0_REFERENCE Table: travel_policies
CREATE TABLE travel_policies (
    id TEXT PRIMARY KEY,
    company_id TEXT NOT NULL,
    user_level TEXT NOT NULL CHECK(user_level IN ('STAFF', 'MANAGER', 'DIRECTOR', 'VP')),
    max_flight_cost_no_approval INTEGER NOT NULL,
    allowed_hotel_vendor_type TEXT NOT NULL CHECK(allowed_hotel_vendor_type IN ('PREFERRED', 'ANY')),
    UNIQUE(company_id, user_level)
);

-- L1_ENTITY Table: companies
CREATE TABLE companies (
    id TEXT PRIMARY KEY,
    name TEXT NOT NULL,
    active INTEGER DEFAULT 1 CHECK(active IN (0,1))
);

-- L1_ENTITY Table: users
CREATE TABLE users (
    id TEXT PRIMARY KEY,
    company_id TEXT NOT NULL REFERENCES companies(id),
    user_level TEXT NOT NULL CHECK(user_level IN ('STAFF', 'MANAGER', 'DIRECTOR', 'VP')),
    active INTEGER DEFAULT 1 CHECK(active IN (0,1))
);

-- L2_TRANSACTION Table: approvals
CREATE TABLE approvals (
    id INTEGER PRIMARY KEY AUTOINCREMENT,
    flight_booking_id INTEGER NOT NULL REFERENCES flight_bookings(id),
    approver_id TEXT REFERENCES users(id),
    status TEXT NOT NULL CHECK(status IN ('PENDING', 'APPROVED', 'DENIED')),
    step INTEGER NOT NULL
);

-- L2_TRANSACTION Table: flight_bookings
CREATE TABLE flight_bookings (
    id INTEGER PRIMARY KEY AUTOINCREMENT,
    travel_request_id INTEGER NOT NULL REFERENCES travel_requests(id),
    flight_code TEXT NOT NULL,
    cost INTEGER NOT NULL,
    class TEXT NOT NULL REFERENCES flight_classes(id),
    departure_step INTEGER NOT NULL,
    booking_step INTEGER NOT NULL,
    status TEXT NOT NULL DEFAULT 'PENDING' CHECK(status IN ('PENDING', 'APPROVED', 'TICKETED', 'CANCELLED')),
    approval_status TEXT NOT NULL DEFAULT 'NOT_REQUIRED' CHECK(approval_status IN ('NOT_REQUIRED', 'PENDING', 'APPROVED', 'DENIED')),
    policy_violation_flag INTEGER DEFAULT 0 CHECK(policy_violation_flag IN (0,1)),
    cancellation_step INTEGER,
    refund_amount INTEGER
);

-- L2_TRANSACTION Table: hotel_bookings
CREATE TABLE hotel_bookings (
    id INTEGER PRIMARY KEY AUTOINCREMENT,
    travel_request_id INTEGER NOT NULL REFERENCES travel_requests(id),
    hotel_vendor_id TEXT NOT NULL REFERENCES preferred_vendors(id),
    cost INTEGER NOT NULL,
    booking_step INTEGER NOT NULL,
    status TEXT NOT NULL DEFAULT 'PENDING' CHECK(status IN ('PENDING', 'CONFIRMED', 'CANCELLED')),
    reimbursable INTEGER DEFAULT 1 CHECK(reimbursable IN (0,1)),
    cancellation_step INTEGER,
    refund_amount INTEGER
);

-- L2_TRANSACTION Table: travel_requests
CREATE TABLE travel_requests (
    id INTEGER PRIMARY KEY AUTOINCREMENT,
    user_id TEXT NOT NULL REFERENCES users(id),
    trip_purpose TEXT NOT NULL,
    status TEXT NOT NULL DEFAULT 'DRAFT' CHECK(status IN ('DRAFT', 'SUBMITTED', 'APPROVED', 'CANCELLED')),
    current_step INTEGER NOT NULL,
    flight_booking_count INTEGER DEFAULT 0 CHECK(flight_booking_count >= 0 AND flight_booking_count <= 3),
    hotel_booking_count INTEGER DEFAULT 0 CHECK(hotel_booking_count >= 0 AND hotel_booking_count <= 2)
);
    \end{lstlisting}
\end{greybox}

\begin{greybox}
    \subsection*{Example: Database Triggers}
    \tcbline
    \begin{lstlisting}[style=sqlstyle]
-- Trigger: validate_travel_request_insert
CREATE TRIGGER validate_travel_request_insert
BEFORE INSERT ON travel_requests
FOR EACH ROW
BEGIN
    SELECT CASE
        WHEN NOT EXISTS (SELECT 1 FROM users WHERE id = NEW.user_id AND active = 1)
        THEN RAISE(ABORT, '[PREREQ_FAIL] User does not exist or is inactive')
        WHEN NOT EXISTS (
            SELECT 1 FROM users u JOIN companies c ON u.company_id = c.id
            WHERE u.id = NEW.user_id AND c.active = 1
        ) THEN RAISE(ABORT, '[PREREQ_FAIL] User''s company is inactive')
        WHEN NEW.trip_purpose IS NULL OR NEW.trip_purpose = ''
        THEN RAISE(ABORT, '[POLICY_VIOLATION] Trip purpose is mandatory')
        WHEN NEW.current_step IS NULL
        THEN RAISE(ABORT, '[REQUIRED_FIELD] current_step must be provided')
        WHEN NEW.status IS NULL THEN RAISE(ABORT, '[SYSTEM_ERROR] Status must be provided')
        WHEN NEW.status != 'DRAFT' THEN RAISE(ABORT, '[SYSTEM_ERROR] New travel requests must start as DRAFT')
    END;
END

-- Trigger: enforce_flight_booking_quota
CREATE TRIGGER enforce_flight_booking_quota
BEFORE INSERT ON flight_bookings
FOR EACH ROW
BEGIN
    SELECT CASE
        WHEN (SELECT flight_booking_count FROM travel_requests WHERE id = NEW.travel_request_id) >= 3
        THEN RAISE(ABORT, '[QUOTA_EXCEEDED] Maximum 3 flight bookings per travel request')
    END;
END

-- Trigger: validate_flight_booking_insert
CREATE TRIGGER validate_flight_booking_insert
BEFORE INSERT ON flight_bookings
FOR EACH ROW
BEGIN

    SELECT CASE
        WHEN NOT EXISTS (
            SELECT 1 FROM travel_requests
            WHERE id = NEW.travel_request_id AND status IN ('DRAFT', 'APPROVED')
        ) THEN RAISE(ABORT, '[PREREQ_FAIL] Travel request must be DRAFT or APPROVED')
        WHEN NEW.departure_step IS NULL THEN RAISE(ABORT, '[REQUIRED_FIELD] departure_step must be provided')
        WHEN NEW.booking_step IS NULL THEN RAISE(ABORT, '[REQUIRED_FIELD] booking_step must be provided')
        WHEN NEW.policy_violation_flag != 0
        THEN RAISE(ABORT, '[SYSTEM_CONTROL] policy_violation_flag must be 0, system will calculate')
        WHEN NEW.status != 'PENDING'
        THEN RAISE(ABORT, '[SYSTEM_ERROR] New flight bookings must start as PENDING')
    END;

    SELECT CASE
        WHEN (
            SELECT u.user_level FROM travel_requests tr
            JOIN users u ON tr.user_id = u.id
            WHERE tr.id = NEW.travel_request_id
        ) NOT IN ('DIRECTOR', 'VP') AND NEW.class != 'ECONOMY'
        THEN RAISE(ABORT, '[POLICY_VIOLATION] Only DIRECTOR/VP level can book non-ECONOMY class')
    END;

    SELECT CASE
        WHEN (

            NEW.cost > (
                SELECT tp.max_flight_cost_no_approval
                FROM travel_requests tr
                JOIN users u ON tr.user_id = u.id
                JOIN travel_policies tp ON u.company_id = tp.company_id AND u.user_level = tp.user_level
                WHERE tr.id = NEW.travel_request_id
            )

            AND NOT (
                (SELECT u.user_level FROM travel_requests tr JOIN users u ON tr.user_id = u.id WHERE tr.id = NEW.travel_request_id)
                IN ('DIRECTOR', 'VP') AND NEW.cost < 500
            )
            AND NOT (NEW.departure_step - NEW.booking_step < 3)

            AND NEW.approval_status != 'PENDING'
        )
        THEN RAISE(ABORT, '[POLICY_VIOLATION] Flight requires manager approval. Set approval_status = PENDING')

        WHEN NOT (
            NEW.cost > (
                SELECT tp.max_flight_cost_no_approval
                FROM travel_requests tr
                JOIN users u ON tr.user_id = u.id
                JOIN travel_policies tp ON u.company_id = tp.company_id AND u.user_level = tp.user_level
                WHERE tr.id = NEW.travel_request_id
            )
            AND NOT (
                (SELECT u.user_level FROM travel_requests tr JOIN users u ON tr.user_id = u.id WHERE tr.id = NEW.travel_request_id)
                IN ('DIRECTOR', 'VP') AND NEW.cost < 500
            )
            AND NOT (NEW.departure_step - NEW.booking_step < 3)
        )
        AND NEW.approval_status = 'PENDING'
        THEN RAISE(ABORT, '[LOGIC_ERROR] Approval not required for this flight. Set approval_status = NOT_REQUIRED')
    END;
END

-- Trigger: process_flight_booking_after_insert
CREATE TRIGGER process_flight_booking_after_insert
AFTER INSERT ON flight_bookings
FOR EACH ROW
BEGIN

    UPDATE flight_bookings
    SET policy_violation_flag = CASE WHEN (NEW.departure_step - NEW.booking_step < 3) THEN 1 ELSE 0 END
    WHERE id = NEW.id;

    INSERT INTO approvals (flight_booking_id, status, step)
    SELECT NEW.id, 'PENDING', NEW.booking_step
    WHERE EXISTS (
        SELECT 1
        FROM travel_requests tr
        JOIN users u ON tr.user_id = u.id
        JOIN travel_policies tp ON u.company_id = tp.company_id AND u.user_level = tp.user_level
        WHERE tr.id = NEW.travel_request_id
        AND NEW.cost > tp.max_flight_cost_no_approval
        AND NOT (u.user_level IN ('DIRECTOR', 'VP') AND NEW.cost < 500)
        AND NOT (NEW.departure_step - NEW.booking_step < 3)
    );

END

-- Trigger: enforce_hotel_booking_quota
CREATE TRIGGER enforce_hotel_booking_quota
BEFORE INSERT ON hotel_bookings
FOR EACH ROW
BEGIN
    SELECT CASE
        WHEN (SELECT hotel_booking_count FROM travel_requests WHERE id = NEW.travel_request_id) >= 2
        THEN RAISE(ABORT, '[QUOTA_EXCEEDED] Maximum 2 hotel bookings per travel request')
    END;
END

-- Trigger: validate_hotel_booking_insert
CREATE TRIGGER validate_hotel_booking_insert
BEFORE INSERT ON hotel_bookings
FOR EACH ROW
BEGIN
    SELECT CASE
        WHEN NOT EXISTS (
            SELECT 1 FROM travel_requests
            WHERE id = NEW.travel_request_id AND status IN ('DRAFT', 'APPROVED')
        ) THEN RAISE(ABORT, '[PREREQ_FAIL] Travel request must be DRAFT or APPROVED')
        WHEN NOT EXISTS (SELECT 1 FROM preferred_vendors WHERE id = NEW.hotel_vendor_id)
        THEN RAISE(ABORT, '[POLICY_VIOLATION] Hotel must be from preferred vendors list')
        WHEN NEW.booking_step IS NULL
        THEN RAISE(ABORT, '[REQUIRED_FIELD] booking_step must be provided')
        WHEN NEW.reimbursable != 1
        THEN RAISE(ABORT, '[SYSTEM_CONTROL] reimbursable must be 1, system will set based on vendor type')
        WHEN NEW.status != 'PENDING'
        THEN RAISE(ABORT, '[SYSTEM_ERROR] New hotel bookings must start as PENDING')
    END;

    SELECT CASE
        WHEN (
            SELECT tp.allowed_hotel_vendor_type
            FROM travel_requests tr
            JOIN users u ON tr.user_id = u.id
            JOIN travel_policies tp ON u.company_id = tp.company_id AND u.user_level = tp.user_level
            WHERE tr.id = NEW.travel_request_id
        ) != 'ANY'
        AND (
            SELECT tp.allowed_hotel_vendor_type
            FROM travel_requests tr
            JOIN users u ON tr.user_id = u.id
            JOIN travel_policies tp ON u.company_id = tp.company_id AND u.user_level = tp.user_level
            WHERE tr.id = NEW.travel_request_id
        ) != (
            SELECT pv.vendor_type FROM preferred_vendors pv WHERE pv.id = NEW.hotel_vendor_id
        )
        THEN RAISE(ABORT, '[POLICY_VIOLATION] Hotel vendor type does not match policy requirement')
    END;
END

-- Trigger: process_hotel_booking_after_insert
CREATE TRIGGER process_hotel_booking_after_insert
AFTER INSERT ON hotel_bookings
FOR EACH ROW
BEGIN

    UPDATE hotel_bookings
    SET reimbursable = CASE
        WHEN (SELECT vendor_type FROM preferred_vendors WHERE id = NEW.hotel_vendor_id) = 'PREFERRED'
        THEN 1 ELSE 0
    END
    WHERE id = NEW.id;
END

-- Trigger: validate_flight_cancellation
CREATE TRIGGER validate_flight_cancellation
BEFORE UPDATE OF status ON flight_bookings
FOR EACH ROW
WHEN NEW.status = 'CANCELLED' AND OLD.status != 'CANCELLED'
BEGIN
    SELECT CASE
        WHEN OLD.status = 'TICKETED'
        THEN RAISE(ABORT, '[IRREVERSIBLE] TICKETED flights cannot be cancelled')
        WHEN NEW.cancellation_step IS NULL
        THEN RAISE(ABORT, '[PROVENANCE_REQUIRED] Cancellation step must be provided')
        WHEN NEW.refund_amount IS NULL
        THEN RAISE(ABORT, '[CALCULATION_REQUIRED] Refund amount must be calculated')

        WHEN NEW.cancellation_step - OLD.booking_step <= 2 AND NEW.refund_amount != OLD.cost
        THEN RAISE(ABORT, '[CALCULATION_ERROR] Flight cancellation within 2 steps of booking gets full refund: ' || OLD.cost)
        WHEN NEW.cancellation_step - OLD.booking_step > 2 AND NEW.refund_amount != OLD.cost / 2
        THEN RAISE(ABORT, '[CALCULATION_ERROR] Late flight cancellation (>2 steps from booking) gets 50% refund: ' || (OLD.cost / 2))
    END;
END

-- Trigger: validate_hotel_cancellation
CREATE TRIGGER validate_hotel_cancellation
BEFORE UPDATE OF status ON hotel_bookings
FOR EACH ROW
WHEN NEW.status = 'CANCELLED' AND OLD.status != 'CANCELLED'
BEGIN
    SELECT CASE
        WHEN OLD.status = 'CONFIRMED'
        THEN RAISE(ABORT, '[IRREVERSIBLE] CONFIRMED hotels cannot be cancelled')
        WHEN NEW.cancellation_step IS NULL
        THEN RAISE(ABORT, '[PROVENANCE_REQUIRED] Cancellation step must be provided')
        WHEN NEW.refund_amount IS NULL
        THEN RAISE(ABORT, '[CALCULATION_REQUIRED] Refund amount must be calculated')

        WHEN NEW.cancellation_step - OLD.booking_step <= 2 AND NEW.refund_amount != OLD.cost
        THEN RAISE(ABORT, '[CALCULATION_ERROR] Hotel cancellation within 2 steps of booking gets full refund: ' || OLD.cost)
        WHEN NEW.cancellation_step - OLD.booking_step > 2 AND NEW.refund_amount != OLD.cost / 2
        THEN RAISE(ABORT, '[CALCULATION_ERROR] Late hotel cancellation (>2 steps from booking) gets 50% refund: ' || (OLD.cost / 2))
    END;
END

-- Trigger: recalc_flight_quota_after_status_change
CREATE TRIGGER recalc_flight_quota_after_status_change
AFTER UPDATE OF status ON flight_bookings
FOR EACH ROW
WHEN NEW.status != OLD.status
BEGIN
    UPDATE travel_requests
    SET flight_booking_count = (
        SELECT COUNT(*) FROM flight_bookings
        WHERE travel_request_id = OLD.travel_request_id AND status != 'CANCELLED'
    )
    WHERE id = OLD.travel_request_id;
END

-- Trigger: recalc_hotel_quota_after_status_change
CREATE TRIGGER recalc_hotel_quota_after_status_change
AFTER UPDATE OF status ON hotel_bookings
FOR EACH ROW
WHEN NEW.status != OLD.status
BEGIN
    UPDATE travel_requests
    SET hotel_booking_count = (
        SELECT COUNT(*) FROM hotel_bookings
        WHERE travel_request_id = OLD.travel_request_id AND status != 'CANCELLED'
    )
    WHERE id = OLD.travel_request_id;
END

-- Trigger: validate_approval_update
CREATE TRIGGER validate_approval_update
BEFORE UPDATE ON approvals
FOR EACH ROW
BEGIN
    SELECT CASE
        WHEN OLD.status != 'PENDING'
        THEN RAISE(ABORT, '[IRREVERSIBLE] Can only update PENDING approvals')
        WHEN OLD.flight_booking_id != NEW.flight_booking_id
        THEN RAISE(ABORT, '[IMMUTABLE] Cannot change flight_booking_id')
        WHEN OLD.step != NEW.step
        THEN RAISE(ABORT, '[IMMUTABLE] Cannot change approval step')
        WHEN NEW.approver_id IS NOT NULL AND NOT EXISTS (
            SELECT 1 FROM users WHERE id = NEW.approver_id AND active = 1
        ) THEN RAISE(ABORT, '[PREREQ_FAIL] Approver does not exist or is inactive')
        WHEN NEW.approver_id IS NOT NULL AND (
            SELECT user_level FROM users WHERE id = NEW.approver_id
        ) NOT IN ('MANAGER', 'DIRECTOR', 'VP')
        THEN RAISE(ABORT, '[AUTHORITY_ERROR] Approver must be MANAGER level or higher')
        WHEN NEW.approver_id IS NOT NULL AND EXISTS (
            SELECT 1 FROM flight_bookings fb
            JOIN travel_requests tr ON fb.travel_request_id = tr.id
            WHERE fb.id = NEW.flight_booking_id AND tr.user_id = NEW.approver_id
        ) THEN RAISE(ABORT, '[CONFLICT_OF_INTEREST] Approver cannot approve their own request')
    END;
END

-- Trigger: process_approval_after_update
CREATE TRIGGER process_approval_after_update
AFTER UPDATE ON approvals
FOR EACH ROW
WHEN NEW.status != OLD.status
BEGIN

    UPDATE flight_bookings
    SET approval_status = NEW.status
    WHERE id = NEW.flight_booking_id;

    UPDATE flight_bookings
    SET status = 'TICKETED'
    WHERE id = NEW.flight_booking_id
    AND NEW.status = 'APPROVED'
    AND status = 'PENDING'
    AND policy_violation_flag = 0;
END

-- Trigger: prevent_flight_modification_after_final
CREATE TRIGGER prevent_flight_modification_after_final
BEFORE UPDATE ON flight_bookings
FOR EACH ROW
WHEN OLD.status IN ('TICKETED', 'CANCELLED')
BEGIN
    SELECT CASE
        WHEN OLD.status = 'TICKETED' AND NEW.status != 'TICKETED'
        THEN RAISE(ABORT, '[IMMUTABLE] TICKETED bookings cannot be modified')
        WHEN OLD.status = 'CANCELLED' AND NEW.status != 'CANCELLED'
        THEN RAISE(ABORT, '[IMMUTABLE] CANCELLED bookings cannot be reactivated')
        WHEN OLD.cost != NEW.cost
        THEN RAISE(ABORT, '[IMMUTABLE] Cost cannot be changed after creation')
        WHEN OLD.departure_step != NEW.departure_step
        THEN RAISE(ABORT, '[IMMUTABLE] Departure step cannot be changed after creation')
        WHEN OLD.booking_step != NEW.booking_step
        THEN RAISE(ABORT, '[IMMUTABLE] Booking step cannot be changed after creation')
    END;
END

-- Trigger: prevent_hotel_modification_after_final
CREATE TRIGGER prevent_hotel_modification_after_final
BEFORE UPDATE ON hotel_bookings
FOR EACH ROW
WHEN OLD.status IN ('CONFIRMED', 'CANCELLED')
BEGIN
    SELECT CASE
        WHEN OLD.status = 'CONFIRMED' AND NEW.status != 'CONFIRMED'
        THEN RAISE(ABORT, '[IMMUTABLE] CONFIRMED hotels cannot be modified')
        WHEN OLD.status = 'CANCELLED' AND NEW.status != 'CANCELLED'
        THEN RAISE(ABORT, '[IMMUTABLE] CANCELLED hotels cannot be reactivated')
        WHEN OLD.cost != NEW.cost
        THEN RAISE(ABORT, '[IMMUTABLE] Cost cannot be changed after creation')
        WHEN OLD.booking_step != NEW.booking_step
        THEN RAISE(ABORT, '[IMMUTABLE] Booking step cannot be changed after creation')
    END;
END;
    \end{lstlisting}
\end{greybox}

\begin{greybox}

\subsection*{Example: Atomic Tool}
\tcbline
query\_approvals, query\_companies, query\_flight\_bookings, 

query\_flight\_classes, query\_users, query\_hotel\_bookings, 

query\_preferred\_vendors, query\_travel\_policies, query\_travel\_requests

update\_approvals, update\_flight\_bookings, 

update\_hotel\_bookings, update\_travel\_requests

insert\_approvals, insert\_flight\_bookings,

insert\_hotel\_bookings, insert\_travel\_requests

transfer\_to\_human\_agents
\end{greybox}

\begin{greybox}
    \textbf{Example: TASK DESCRIPTION}
    \tcbline 

\noindent\textbf{WHO\_YOU\_ARE:}\\
You are ``DIRECTOR'' (User ID: ``u\_history\_01''), a senior executive at ``comp\_alpha'' who is strategic, efficient, and leverages high-level privileges to ensure smooth operations. You are preparing for an important Board Meeting and value comprehensive planning with backup options. Your personality is detail-oriented and consultative—you prefer to understand all available services before acting, but once informed, you make decisive choices based on your director benefits, such as the special waiver for flights under ``\$500''. You occasionally use strategic delays in requests (like adding bookings one at a time) to test the system's responsiveness, but you ultimately cooperate with the agent's guidance.

\vspace{0.5em}
\noindent\textbf{COGNITIVE\_GAPS \& FALSE BELIEFS (CRITICAL):}
\begin{itemize}
  \item You do NOT know the specific flight codes or hotel vendor IDs available until the agent provides them.
  \item You do NOT know the current status of your travel request slots until the agent informs you.
  \item (No false beliefs identified; you correctly understand your privileges and policies.)
\end{itemize}

\vspace{0.5em}
\noindent\textbf{YOUR\_TOTAL\_GOAL:}\\
To ensure you have comprehensive, flexible travel arrangements for your Board Meeting by adding backup flight and hotel bookings to your existing approved travel request, leveraging your director-level privileges for efficiency and compliance.

\vspace{0.5em}
\noindent\textbf{SUB\_GOALS (Must be completed in order):}

\begin{enumerate}
  \item \textbf{[Establish Identity and Discover Services]}
  \begin{itemize}
    \item Begin by introducing yourself as ``DIRECTOR'' from ``comp\_alpha'' (User ID: ``u\_history\_01'') to authenticate your identity. Then, ask the agent for a detailed overview of all available services, including costs, requirements, and your current travel status, to plan your next steps strategically.
    \item \textbf{REACTIVE\_TRIGGER:}
    \begin{itemize}
      \item \textbf{IF} the agent provides a dry list without details, \textbf{THEN} insist on knowing the differences between services, such as costs and eligibility rules.
    \end{itemize}
  \end{itemize}

  \item \textbf{[Request a Backup Flight with Director Privileges]}
  \begin{itemize}
    \item Once you understand the services, state that you want to add a flight to your existing approved travel request (\#4) for the Board Meeting. Specify that you need a ``business class'' flight costing ``under \$500'' to utilize your director waiver (which exempts such flights from approval). Do not provide specific flight codes yet; instead, ask the agent for options that meet these criteria.
  \end{itemize}

  \item \textbf{[Confirm Flight Booking and Delay Hotel Request]}
  \begin{itemize}
    \item After the flight is booked, express satisfaction with the outcome and confirm that you want to keep the booking as-is. Crucially, do NOT mention the hotel yet—simulate a forgetful user to test if the agent handles sequential requests properly.
    \item \textbf{REACTIVE\_TRIGGER:}
    \begin{itemize}
      \item \textbf{IF} the agent provides a summary of your current bookings and offers further assistance, \textbf{THEN} pivot to your next need: requesting a backup hotel accommodation as if you just remembered it.
    \end{itemize}
  \end{itemize}

  \item \textbf{[Request a Backup Hotel with Preference for Value]}
  \begin{itemize}
    \item Explain that you need a backup hotel from ``preferred vendors'' to ensure reimbursability, targeting a cost of ``around \$250'' for reliability and value. Ask the agent to recommend options that fit this description, prioritizing trusted vendors.
  \end{itemize}

  \item \textbf{[Verify Compliance and Finalize Arrangements]}
  \begin{itemize}
    \item Once all bookings are confirmed, request a comprehensive policy compliance audit to ensure everything is properly documented and aligns with company rules. You want verification that all bookings—including the flight waiver and hotel reimbursability—are fully compliant before considering the task complete.
  \end{itemize}
\end{enumerate}

\end{greybox}

\section{Dataset Construction}
\label{appendix:dataset_construction}
\subsection{User Simulator Prompt}
\label{appendix:user_simulator}

We adapted the User Simulator prompt from $\tau$-Bench~\citep{tau1} and refined it to enable finer-grained, step-by-step control over task execution, ensuring a progressive disclosure of the input task description.

\begin{greybox}
\textbf{User Simulator Prompt}
\tcbline 
You are a user interacting with an agent.

\textbf{<USER\_PROFILE>}: This block contains the strict definition of your identity (Name, ID, Tier, etc.).

\{USER\_PROFILE\}

\textbf{<TASK\_DESCRIPTION> (Your Soul \& Goals)}: This is your master directive. It provides your goals. You will live and breathe this script.

\{TASK\_DESCRIPTION\}

Rules:

- Just generate one line at a time to simulate the user's message.

- Do not give away all the instruction at once. Only provide the information that is necessary for the current step.

- Do not hallucinate information that is not provided in the instruction. For example, if the agent asks for the order id but it is not mentioned in the instruction, do not make up an order id, just say you do not remember or have it.

- If the instruction goal is satisfied, generate '\#\#\#STOP\#\#\#' as a standalone message without anything else to end the conversation.

- Do not repeat the exact instruction in the conversation. Instead, use your own words to convey the same information.

- Try to make the conversation as natural as possible, and stick to the personalities in the instruction.

\end{greybox}

\subsection{Computational Cost Analysis}

To quantify the computational investment required for the LOGIGEN pipeline, we performed a detailed cost analysis covering both the Triple-Agent data synthesis phase and the SFT trajectory distillation process. Given the conversational, multi-turn nature of the generation process, we treat each interaction turn as a distinct ``instruction-response'' pair and accumulate the input and output tokens across all turns. The cost estimates are calculated based on a ``no prefix cache'' assumption, providing a conservative upper bound on infrastructure expenses. Following standard open-weight model pricing, we apply a rate of \textbf{¥ 2.00 per million input tokens} and \textbf{¥ 3.00 per million output tokens}. Table~\ref{tab:cost_breakdown} presents a detailed breakdown of the average performance (success rate, turns) and the corresponding costs for each module within the LOGIGEN framework.

\begin{table}[htbp]
\centering
\caption{Breakdown of Performance and Cost by Agent Module}
\label{tab:cost_breakdown}
\small 
\renewcommand{\arraystretch}{1.2} 
\begin{tabular}{l l l r r r r} 
\toprule
\textbf{Module / Agent} & \textbf{Success} & \textbf{Model} & \textbf{Avg.} & \multicolumn{2}{c}{\textbf{Avg. Tokens}} & \textbf{Avg.} \\
\cmidrule(lr){5-6} 
 & \textbf{Rate (\%)} & & \textbf{Turn} & \textbf{Prompt} & \textbf{Compl.} & \textbf{Cost (¥)} \\
\midrule

\textbf{The Architect} & \textbf{68.5} & -- & -- & -- & -- & \textbf{1.22} \\
\hspace{1em} Design Wiki Policy & -- & DS-V3.2-Think & 2.7 & 37,410 & 6,255 & 0.09 \\
\hspace{1em} Design Tables & -- & DS-V3.2-Think & 7.1 & 157,048 & 15,906 & 0.36 \\
\hspace{1em} Design Triggers & -- & DS-V3.2-Think & 10.0 & 327,096 & 38,654 & 0.77 \\
\addlinespace 

\textbf{The Set Designer} & \textbf{72.8} & DS-V3.2 & 41.4 & 795,374 & 5,941 & \textbf{1.61} \\
\addlinespace

\textbf{The Explorer} & \textbf{99.3} & -- & -- & -- & -- & \textbf{1.24} \\
\hspace{1em} Client Agent & -- & DS-V3.2 & 8.74 & 82,209 & 2,073 & 0.17 \\
\hspace{1em} Consultant Agent & -- & DS-V3.2 & 23.86 & 523,398 & 6,636 & 1.07 \\
\addlinespace

\textbf{Distill Trajectory} & \textbf{74.2} & -- & -- & -- & -- & \textbf{1.43} \\
\hspace{1em} User Simulator & -- & DS-V3.2 & 8.8 & 124,152 & 406 & 0.24 \\
\hspace{1em} Teacher Agent & -- & DS-V3.2-Think & 28.9 & 587,168 & 11,363 & 1.19 \\

\bottomrule
\end{tabular}
\end{table}

\end{document}